\documentclass[journal]{IEEEtran}
\usepackage{amsmath}
\usepackage{amssymb}
\usepackage[noend]{algpseudocode}
\usepackage{algorithmicx,algorithm}
\usepackage{url}
\usepackage{graphicx}
\usepackage{bigstrut,multirow,rotating}
\usepackage{threeparttable}
\usepackage{booktabs}
\usepackage{array}
\usepackage{bm}
\usepackage[table]{xcolor}
\newcommand{\tabincell}[2]{\begin{tabular}{@{}#1@{}}#2\end{tabular}}

\ifCLASSINFOpdf
\else
\fi
\begin{document}
\title{Generalization-aware Remote Sensing Change Detection via Domain-agnostic Learning}

\author{Qi Zang,
        Shuang Wang, \IEEEmembership{Senior Member, IEEE,}
        Dong Zhao, 
        Dou Quan, \IEEEmembership{Member, IEEE,}
        Yang Hu,
        Zhi Yang,
        and Licheng Jiao, \IEEEmembership{Fellow, IEEE}
\thanks{Manuscript received 1 September 2024; revised 13 January 2025 and 11 March 2025; accepted 12 March 2025. This work was supported by the National Natural Science Foundation of China under Grant Nos. 62271377, the National Key Research and Development Program of China under Grant Nos. 2021ZD0110400, 2021ZD0110404, the Key Research and Development Program of Shannxi (Program Nos. 2023YBGY244, 2023QCYLL28, 2024GX-ZDCYL-02-08, 2024GX-ZDCYL-02-17),  the Key Scientific Technological Innovation Research Project by Ministry of Education, the Joint Funds of the National Natural Science Foundation of China (U22B2054).}
\thanks{Qi Zang, Shuang Wang, Dong Zhao, Dou Quan, Yang Hu, and Licheng Jiao are with the Key Laboratory of Intelligent Perception and Image Understanding of Ministry of Education, School of Artificial Intelligence, Xidian University, Xi'an 710071, China (e-mail: shwang@mail.xidian.edu.cn).\emph{(Corresponding author: Shuang Wang).}}
\thanks{Zhi Yang is with the DFH Satellite Co., Ltd, CAST, Beijing 100080, China.}}

%
%

\markboth{Journal of \LaTeX\ Class Files,~Vol.~14, No.~8, July~2024}%
{Shell \MakeLowercase{\textit{et al.}}: Bare Demo of IEEEtran.cls for IEEE Journals}
%


\maketitle
\begin{abstract}
Change detection has essential significance for the region's development, in which pseudo-changes between bitemporal images induced by imaging environmental factors are key challenges. Existing transformation-based methods regard pseudo-changes as a kind of style shift and alleviate it by transforming bitemporal images into the same style using generative adversarial networks (GANs). However, their efforts are limited by two drawbacks: 1) Transformed images suffer from distortion that reduces feature discrimination. 2) Alignment hampers the model from learning domain-agnostic representations that degrades performance on scenes with domain shifts from the training data. Therefore, oriented from pseudo-changes caused by style differences, we present a generalizable domain-agnostic difference learning network (DonaNet). For the drawback 1), we argue for local-level statistics as style proxies to assist against domain shifts. For the drawback 2), DonaNet learns domain-agnostic representations by removing domain-specific style of encoded features and highlighting the class characteristics of objects. In the removal, we propose a domain difference removal module to reduce feature variance while preserving discriminative properties and propose its enhanced version to provide possibilities for eliminating more style by decorrelating the correlation between features. 
In the highlighting, we propose a cross-temporal generalization learning strategy to imitate latent domain shifts, thus enabling the model to extract feature representations more robust to shifts actively.
Extensive experiments conducted on three public datasets demonstrate that DonaNet outperforms existing state-of-the-art methods with a smaller model size and is more robust to domain shift.
\end{abstract}

\IEEEpeerreviewmaketitle

\section{Introduction}
\IEEEPARstart{R}{emote} sensing change detection (CD) is the process of identifying differences in the state of an object or phenomenon acquired in the same geographic area but at different phases \cite{bruzzone2012novel}. Much more attention~\cite{zang2024ChangeDiff} has been recently paid to the impact of the earth's environmental change on human survival. This urges the government and private enterprises to focus on the development of CD, which has the potential to be used for various applications, such as land use survey \cite{feranec2007corine}, urban expansion \cite{kadhim2016advances}, natural disaster monitoring and assessment \cite{koltunov2007early}, and ecological environment detection \cite{chen2013multi}.

Traditional CD methods typically employ clustering or threshold segmentation to generate difference result maps to identify changed and unchanged regions. The detection accuracy of such methods depends heavily on the quality of manually extracted features, which are usually inefficient or even inaccurate for CD due to limited and nonadaptive feature extraction methods. By contrast, deep learning (DL) represented by the convolutional neural network (CNN) generally shows more powerful feature extraction capabilities than traditional ones \cite{chen2021enhanced,yang2022continual,wang2022cluster,zhao2023towards,
wang2023select,zhao2023learning,zhao2024stable}. They are now widely applied in various remote sensing image processing tasks and achieve significant success \cite{nie2022mign,zhang2024frequency,zhu2022target,zhao2025connectivity,zang2024joint,zang2024generalized}. In CD, the multiple high-level features extracted by CNN contain rich nonlinear representations in spectral space, significantly enhancing CD performance and contributing to its tremendous progress in recent years \cite{lei2019landslide,peng2019end}.

\begin{figure}[t]
    \setlength{\abovecaptionskip}{-0.35cm}
    \begin{center}
    \centering 
    \includegraphics[width=0.49\textwidth, height=0.24\textwidth]{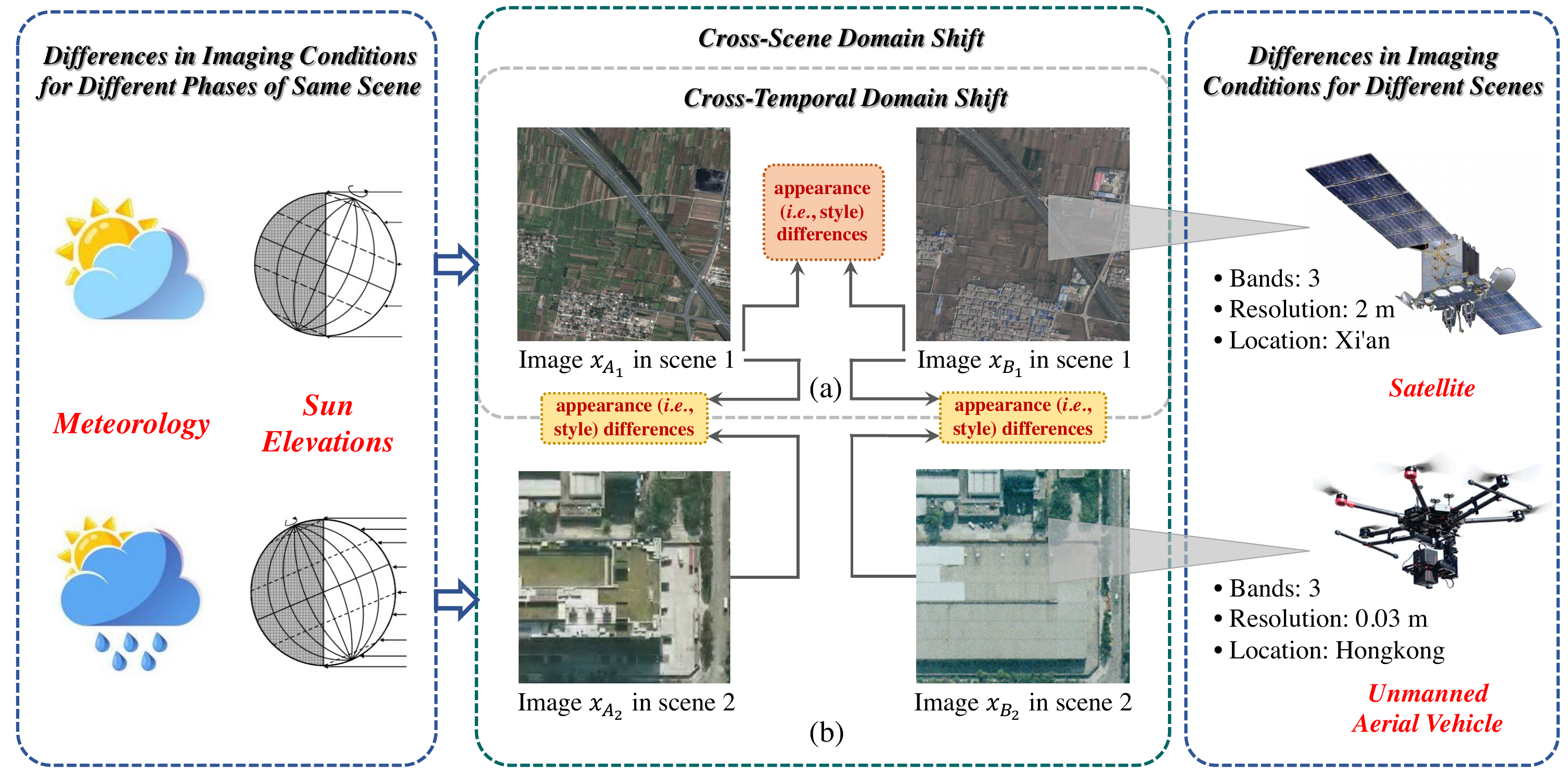} 
    \end{center}
    \caption{Example of domain shift for two types: (a) Cross-temporal domain shifts within a bitemporal image pair. (b) Cross-scene domain shifts between bitemporal image pairs acquired from different scenes.}
    \label{fig16}
\end{figure}

\begin{figure*}[t] 
    \setlength{\abovecaptionskip}{-0.40cm}
    \begin{center}
    \centering 
    \includegraphics[width=1.0\textwidth, height=0.22\textwidth]{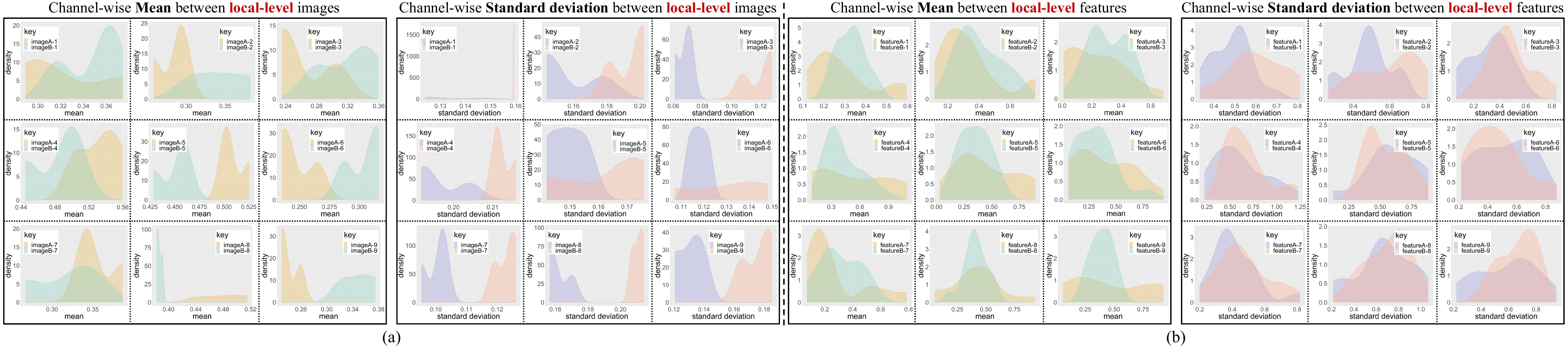} 
    \end{center}
    \caption{(a) Visualization of channel-wise means and standard deviations between local-level bitemporal images. (b) Visualization of channel-wise means and standard deviations between local-level bitemporal features.}
    \label{fig1}
\end{figure*}

Nevertheless, challenges persist in handling pseudo-changes between bitemporal images \cite{hussain2013change,zheng2021clnet}. This issue is induced by variations in acquisition conditions (sun elevation, season, or sensor) that lead to complex object presentation in the scene (spectral behavior is different in the two phases if their semantics do not change), that is, the radiation properties of the same semantics are not similar. These factors significantly affect detection accuracy, as changes in object properties (presence/absence, orientation, size, shape, color, semantic identity) are the sole focus of interest in CD \cite{bruzzone2012novel,liu2014hierarchical,liu2015sequential}. To alleviate pseudo-changes, most methods enhance the discriminability of deep features by improving the network structure \cite{chen2022fccdn,zhang2023global,liu2023attention} or designing various attention mechanisms \cite{chen2020dasnet,zhou2022spatial}. However, these efforts are constrained by a lack of focus on the fundamental cause of pseudo-changes — the disagreement between sensor-measured values and the object's spectral reflectance and radiance \cite{du2002radiometric,teillet2001radiometric}, \emph{i.e.}, spectral shift. Several existing transformation-based methods regard this spectral shift as essentially analogous to a visual appearance or style differences, also known as domain shift \cite{hsu2020progressive} (See Fig. \ref{fig16}(a)). They alleviate this issue mainly by leveraging generative adversarial networks (GANs) to transform bitemporal image styles into the same one \cite{fang2019dual,fang2021content,liu2022end}. 

However, there are two drawbacks in these transformation-based methods: (1) Transformed images frequently suffer distortion from artifacts, primarily attributable to the complexities of optimizing GANs. This phenomenon leads to a decrease in the discriminability of encoded features. (2) The idea of aligning image styles hampers the model from learning domain-agnostic representations due to the involvement of style factors. This degrades the performance of directly generalizing a model tailored for one specific scene (dataset) to another because there is also cross-scene domain shift between different scenes (See Fig. \ref{fig16}(b)).

To tackle the above two drawbacks, this paper follows the insight that pseudo-changes are caused by style differences and develops a generalizable DL network for CD, namely \textbf{d}omain-agn\textbf{o}stic differe\textbf{n}ce le\textbf{a}rning network (DonaNet). DonaNet aims to avoid overfitting to the training domain by forcing the model to learn domain-agnostic difference feature representations. Based on the identified challenges, \emph{i.e.}, how to ensure feature discriminability and how to enable the model generalizable, we proceed as follows:

(1) For the former, we strive to explore a proxy for style information to assist against domain shifts. Recent studies have revealed that the channel-wise mean and standard deviation of images/features carry style \cite{li2021feature,li2022uncertainty}. Whereas, this global representation is overly coarse-grained for CD, undermining crucial semantic details. To explore the fine-grained proxy, given all the data of SVCD dataset \cite{lebedev2018change} where the changed pixels are masked, we divide them into equal-sized local regions and visualize the average of these statistics computed from local-level bitemporal images/features, as shown in Fig. \ref{fig1}. For any given local region, the degree of overlap between the two colors represents the style difference between two images/features. In each local region, we observe that the style shifts vary across the channels. Similarly, when comparing multiple local regions, the degree of overlap between the two colors varies from region to region. Thus, we view channel-wise mean and standard deviation of local-level images/features as styles. Note that unlike methods for other tasks that can omit channel-wise computation \cite{zhao2022source}, we maintain per-channel representation to meet the requirement for discriminative pixel-level comparison in change detection.

(2) For the latter, recent methods for basic vision tasks have explored strategies for cross-domain object recognition \cite{zhang2023learning,zhang2023fine,zhang2024video,liao2024calibration}. This paper tackles difference recognition by treating each pixel as an independent instance to learn its intrinsic content across different styles. Given that local-level statistics can proxy for style, we propose two approaches from different perspectives. \emph{1) Domain Difference Removal Module (DDR).} A straightforward way is to remove the domain-specific style in the encoded difference features. Therefore, we combine regional instance normalization with batch normalization in CNN to remove style and preserve the discriminability of features, namely \emph{global-to-local normalization}.
Nevertheless, normalization does not consider the correlations between features, which may carry redundant style.
Thus, we introduce the whitening transformation to propose an enhanced version of \emph{global-to-local normalization}, \emph{global-to-local whitening}, to such that the features are decorrelated, enhancing the generalization ability of feature representations.
\emph{2) Cross-Temporal Generalization Learning (CTGL).} 
A model that actively extracts essential class characteristics free from style interference can be motivated to identify real changed objects.
Since the essential characteristics of an object are carried by high-level class information in the image, it is reasonable to assume that the model can better extract characteristics of  objects robust to shifts by learning more class information. 
Thus, we perform cross-temporal style transformation with multiple modes on image pairs, which provides appropriate and meaningful style variation for samples by simulating various potential shifts. Further, to eliminate the difference between the network predictions for the pre- and post-transformed samples, we add an explicit constraint to align the predictions in the output space.
In summary, our contribution in this paper has fivefold:
\begin{itemize} 
\item
We propose to decouple style from the image to assist representation learning and propose local-level statistics to fine-grained proxy for style.
\item 
We propose for the first time a generalizable DonaNet to force the model to learn domain-agnostic difference features, thus reducing overdetection of pseudo-changes and improving the generalization ability of the model.
\item 
We propose a DDR module to remove domain-specific style in encoded features while preserving discrimination and propose its enhanced version to provide possibilities for eliminating more style. 
\item 
We propose a CTGL strategy to highlight the class characteristics of objects so that the model extract representations more robust to shifts.
\item 
Extensive experiments show that DonaNet can better cope with cross-temporal/scene domain shifts than existing methods. DonaNet is also easily incorporated into existing models to boost their generalization performance.
\end{itemize}

\section{Related Work}
\subsection{Deep Learning-Based Change Detection}
In recent years, due to the powerful feature representation ability of deep learning (DL), many scholars have introduced the deep convolutional neural network (DCNN) into the change detection task in remote sensing images. DL-based methods can be categorized as two mainstreams: patch-based and image-based methods. 

In the patch-based methods, pixel patches are constructed from raw image pairs or difference images, which are input into a DCNN model to learn the changing relationship of center pixels. Gong \emph{et} \emph{al.} \cite{gong2017superpixel} use superpixel segmentation to generate compact superpixel images and then utilize the textural, spectral, and spatial features between superpixels for change detection by a deep belief network. 
Next, a sparse denoising autoencoder is also used to learn semantic differences between bitemporal patches \cite{lei2019multiscale}. 
Some subsequent studies \cite{arabi2018optical,dong2018local} gradually adopt the siamese DCNN structure due to its efficient feature fusion and representation ability. Besides, \cite{wiratama2018dual} propose a double-density convolutional network. To solve the problem that time-series information cannot be extracted, \cite{mou2018learning} introduce LSTM to obtain the time-series features of the data for detecting changing regions. However, the above methods are insufficient to obtain satisfactory results for the fine-grained change detection. The patch size affects the receptive field of the model, such a small receptive field will lead to the lack of contextual semantics, which is not conducive to the improvement of the model detection ability.

Benefiting from the good performance of the fully convolutional network (FCN) \cite{long2015fully} on the segmentation task, many researchers apply it to the change detection \cite{daudt2018high,lei2019landslide,peng2019end,daudt2018fully,chen2020spatial,zhang2021escnet}. Daudt and Lei \emph{et} \emph{al.} \cite{daudt2018high,lei2019landslide} first introduce the U-Net-based FCN. Peng \emph{et} \emph{al.} \cite{peng2019end} propose an end-to-end approach based on the improved UNet$++$ and employ a multi-side output fusion strategy to combine change maps from different semantic levels. Daudt \emph{et} \emph{al.} \cite{daudt2018fully} propose three fully convolutional siamese structures with skip connections, FC-EF, FC-Siam-conc, and FC-Siam-diff, which obtain multi-scale difference maps and decode these maps to calculate the final result. Since then, such network structures have been widely used in the change detection. In contrast to previous methods that do not refer to any useful spatiotemporal relationships, Chen \emph{et} \emph{al.} \cite{chen2020spatial} design a self-attention mechanism to model spatiotemporal relationships and partition the image into multi-scale subregions to capture spatiotemporal relationships at different scales. Subsequently, Zhang \emph{et} \emph{al.} \cite{zhang2021escnet} combine superpixel segmentation with the DCNN model to make the detected changed regions correspond well to object boundaries. Besides, there are several methods focusing on eliminating the style shift between bitemporal images to reduce pseudo-changes, leveraging generative adversarial networks (GANs) to transform bitemporal image styles into the same one \cite{fang2019dual,fang2021content,liu2022end}. However, their efforts are limited as GANs is difficult to train and prone to cause the transformed image distortion. Further, style alignment does not enable the model to be deployed well on data with domain shifts from the training data. In this paper, oriented from pseudo-changes caused by style differences, we explore a proxy for style information to assist against domain shifts and focus the model on learning domain-agnostic representations to achieve generalization to other data. 
\vspace{-0.2cm}

\subsection{Distribution Alignment in Remote Sensing}
The domain adaptation aims to narrow the domain shift between the labeled source and unlabeled target data. Distribution alignment is a mainstream technique to reduce the distance between two distributions (source and target)\cite{pu2020dual}. In remote sensing, distribution alignment is also concerned to solve various tasks such as scene classification \cite{zhou2018deep,liu2020class} and land cover mapping \cite{yan2019triplet,ji2020generative}. The above tasks exploit distribution alignment to reduce appearance differences between labeled training and unlabeled test data from different times or places, \emph{i.e.}, domain shifts\cite{   pu2023memorizing}. Among them, the domain shifts between the two data are usually narrowed in the two spaces for adaptation: image and feature spaces. 

In the former way, scholars reduce style discrepancy between different domains via the style transfer \cite{benjdira2019unsupervised,tang2020srda,tasar2020colormapgan}. Benjdira \emph{et} \emph{al.} \cite{benjdira2019unsupervised} first apply Cycle-GAN \cite{zhu2017unpaired} to make source and target domains have similar visual styles in the image space. Tasar \emph{et} \emph{al.} \cite{tasar2020colormapgan} simplify the above optimization of GAN and achieve a more stable transformation to make the source and target domains have similar spectral distributions. In the latter way, scholars mainly reduce the feature distribution discrepancy by matching statistical moments \cite{borgwardt2006integrating,sun2016return, pu2023dynamic} or adversarial training \cite{tzeng2017adversarial}. For statistical moment matching, distributions from different domains are often approximated to common distributions to match statistics or use kernel techniques to compute higher-order moments in high-dimensional spaces. For adversarial training, the similarity of the distribution is supervised by the discriminator, and the discriminator and generator are mutually trained by a max-min game. However, these methods are computationally intensive and are not easily nested directly into network designs. Also, they may not be suitable for the change detection, as they may be less discriminative for changing regions while narrowing the domain gap. In contrast, our method tailors plug-and-play modules with small computational cost to the siamese architecture of change detection to maintain feature discriminability while eliminating domain shifts. Besides, in the learning of our method, the model is guided to focus on extracting essential semantics rather than aligning the distributions involved in style information, so that it has stronger generalization.

\begin{figure*}[t] 
    \setlength{\abovecaptionskip}{-0.25cm}
    \begin{center}
    \centering 
    \includegraphics[width=0.79\textwidth, height=0.43\textwidth]{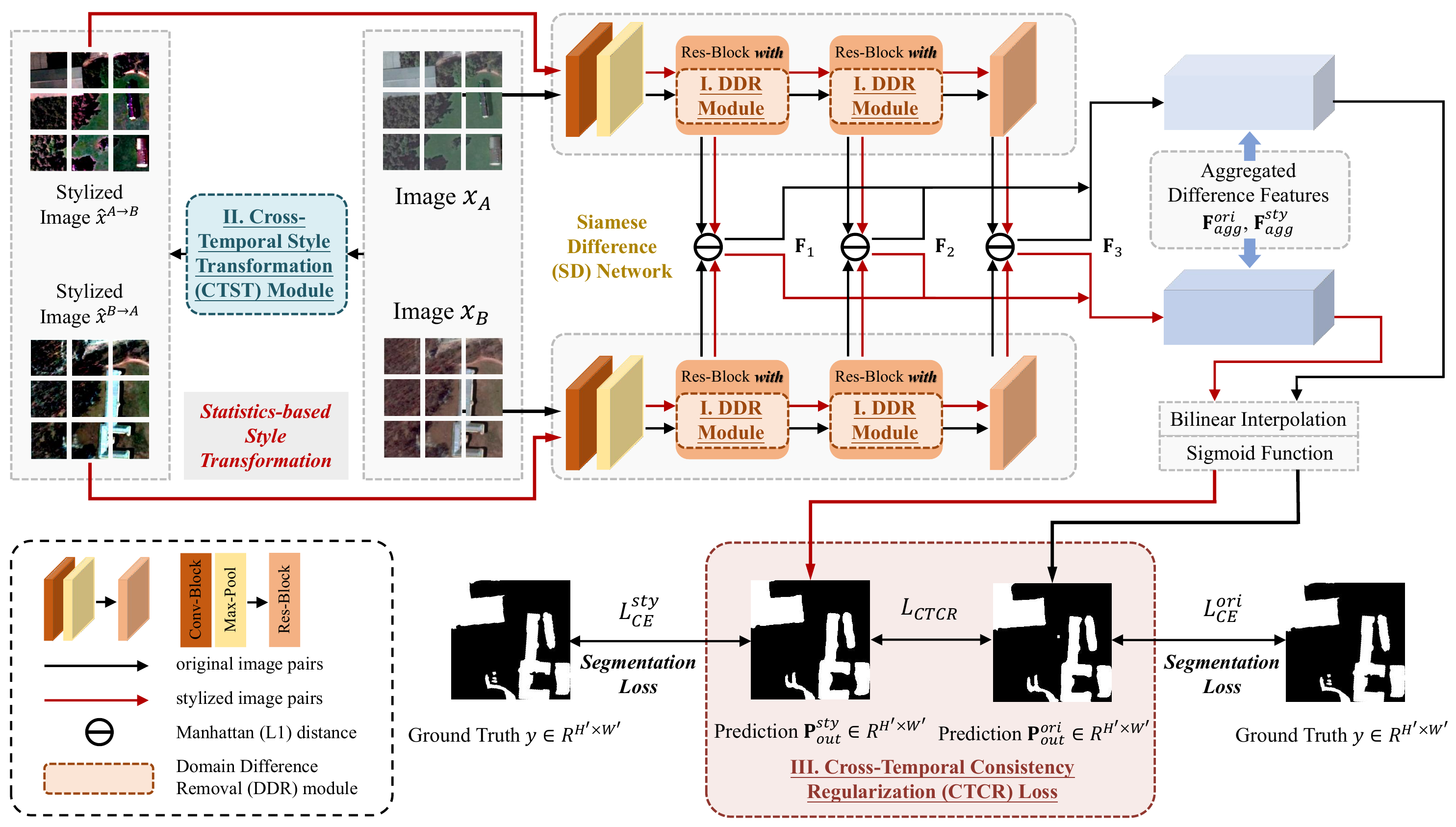} 
    \end{center}
    \caption{An overview of the proposed \textbf{DonaNet}. A pair of images ($x_A$, $x_B$) are first fed into the siamese difference (SD) network. The domain-specific style information in the extracted features is eliminated by the embedded domain difference removal (DDR) module. Then the Manhattan distance is calculated between the features of the two images to obtain multi-level difference features $\{\textbf F_i\}_{i=1}^{3}$. These features are aggregated to generate aggregated features $\textbf F_{agg}$. The cross-temporal style transformation (CTST) module produces stylized image pairs ($\hat x_{}^{A\to B}$, $\hat x_{}^{B\to A}$) via a statistics-based style transformation. The generated stylized image pair is then also fed into the SD network. The obtained result $\textbf P_{out}^{sty}$ is aligned with the result $\textbf P_{out}^{ori}$ of the original image pair in the output space by a cross-temporal consistency regularization (CTCR) loss $\mathcal L_{CTCR}$.} 
    \label{fig2}
\end{figure*}

\section{Proposed Method} \label{3}
\subsection{Overall Framework}
We develop a domain-agnostic difference learning network (DonaNet), forcing the model to learn domain-agnostic difference representations towards actually changed pixels.
The overall framework is shown in Fig. \ref{fig2}. The proposed network mainly consists of four components: siamese difference (SD) network, domain difference removal (DDR) module, cross-temporal style transformation (CTST) module, and cross-temporal consistency regularization (CTCR) loss. 

As shown in Fig. \ref{fig2}, a pair of bitemporal images $x_A$ and $x_B$ from the same area are fed into the 
SD network to extract multi-scale difference feature representations $\textbf F_1$, $\textbf F_2$, and $\textbf F_3$.
The DDR module is a plug-and-play submodule embedded in the SD network, which is designed to eliminate domain-specific style information while maintaining the discriminability of different temporal data. Then, the three-level encoded difference features $\{{\textbf F}_i\}_{i=1}^{3}$ are concatenated to obtain aggregated features $\textbf F_{agg}$. 
The aggregated difference features are resized to the original input size by the bilinear interpolation layer and activated by the sigmoid function to estimate the final output probability score map $\textbf P_{out}$. The map $\textbf P_{out}$ is utilized in a supervised loss function to optimize the SD network. 
Subsequently, the CTST module produces image pairs with diverse styles via statistics-based style transformation and is also used to train the SD network. A consensus is reached between the predictions of the stylized and original image pairs by the CTCR loss, thereby forcing the model to learn domain-agnostic representations.

\subsection{Siamese Difference Network}
The SD network is a two-branch change detection network. The adopted siamese structures separately process bitemporal image pairs by identical branches of the network with shared structure and parameters. With the siamese structure, the feature encoders for each branch are built based on well-known ResNet-18 \cite{he2016deep}. To simplify the model structure and reduce the calculation, the last residual block in ResNet-18 is removed, and only three remain. Fig. \ref{fig3}(a) illustrates the detailed structure of each residual block, containing two convolutional layers with the kernel size of $3 \times 3$ and two batch normalization (BN) layers. The added BN layer prevents the gradient from vanishing, controls the gradient explosion, and speeds up the training and convergence of the network. Then, the input of the residual block itself is added to the output via a skip connection, avoiding the problem of performance degrading as the network depth increases.
For a pair of bitemporal images ($x_A$, $x_B$), the difference features encoded at the three levels $\{\textbf F_i\}_{i=1}^{3}$ can be formulated as,
\begin{equation}
\begin{aligned}
\textbf F_i = \textbf f_{A}^{i}(x_A)\ominus \textbf f_{B}^{i}(x_B),    \label{eq1}
\end{aligned}
\end{equation}
where $\textbf f_{A}^{i}$ and $\textbf f_{B}^{i}$ are the feature maps encoded by the $i$-th residual block of the two branches, respectively. Also, the channel numbers of the three-level difference features are 64, 128, and 256 in turn.

\begin{figure}[t]  
    \setlength{\abovecaptionskip}{-0.35cm}
    \begin{center}
    \centering 
    \includegraphics[width=0.35\textwidth, height=0.27\textwidth]{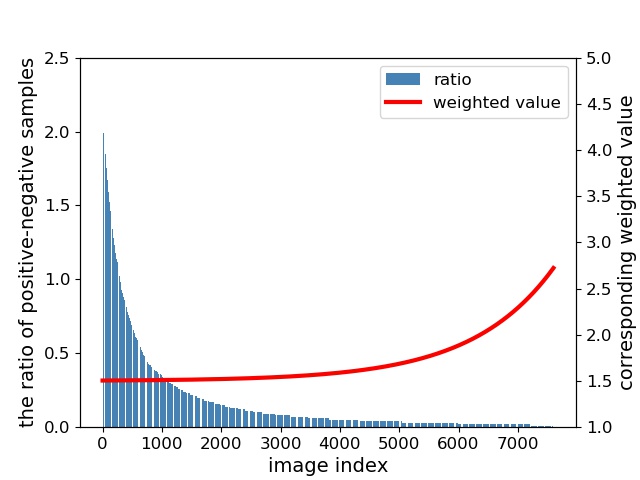} 
    \end{center}
    \caption{The trend of the assigned weights with the corresponding positive-negative sample ratios. 8000 images from the SVCD dataset are randomly sampled as examples.} 
    \label{fig14}
\end{figure}

\begin{figure}[t]  
    \setlength{\abovecaptionskip}{-0.3cm}
    \begin{center}
    \centering 
    \includegraphics[width=0.34\textwidth, height=0.27\textwidth]{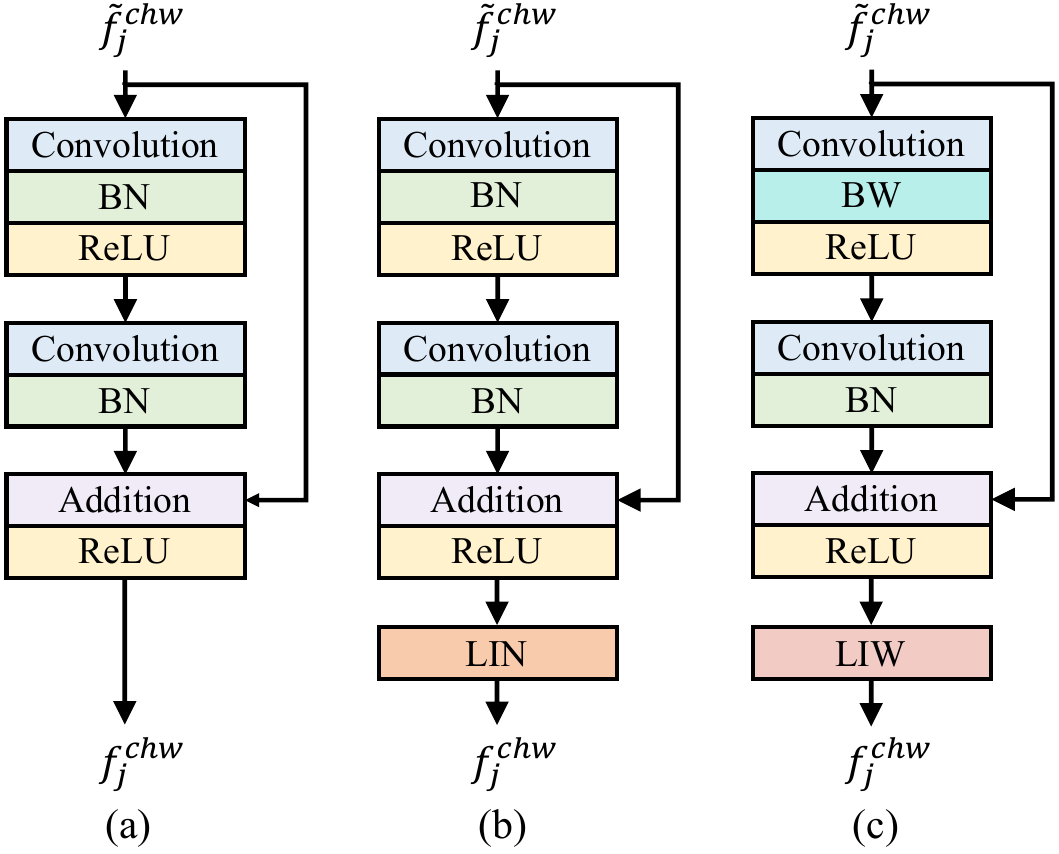} 
    \end{center}
    \caption{Illustration of different structures: (a) The original learnable residual block structure in the SD network. (b) Global-to-local normalization of the domain difference removal (DDR) module. (c) Global-to-local whitening of the domain difference removal (DDR) module.} 
    \label{fig3}
\end{figure}

To make full use of these multi-level features, we aggregate the three-level difference features $\{\textbf F_i\}_{i=1}^{3}$ to generate aggregated features with stronger representation capabilities, so that the changed features at various scales can be taken into account. Specifically, a bilinear interpolation is first employed to resize each difference feature $\textbf F_i$ to the input image. Then, all the resized difference features are concatenated sequentially by the following formula,
\begin{equation}
\textbf F_{agg}=[\textbf F_1||\textbf F_2||\textbf F_3],
\end{equation}
where ``$||$" is the concatenation operation. 
The aggregated difference features $\textbf F_{agg}$ are activated by the sigmoid function to estimate the probability score map $\textbf P_{out}\in \mathbb{R}^{H' \times W'}$, where $H'$ and $W'$ are the height and width of the image. In addition, by calculating the ratio of positive-negative (changed and unchanged) samples/pixels corresponding to different image pairs, we note that the distribution of positive-negative samples is highly imbalanced. This problem of sample imbalance is also focused on by the STANet \cite{chen2020spatial} and ESCNet \cite{zhang2021escnet} works. Thus, a dynamically weighted binary cross-entropy loss function is designed as,
\begin{equation}
\begin{aligned}
\mathcal L_{WCE}&=-\frac{1}{N}\sum_{j=1}^{N}[w_jy_j\log(\textbf P_{out}^{j})+(1-y_j)\log(1-\textbf P_{out}^{j})],\\
w_j&=e_{}^{[\alpha+\beta (n_{pos}^{j}/n_{neg}^{j})]}+\lambda, \label{eq14}
\end{aligned}
\end{equation}
where $N$ is the number of image pairs in a mini-batch, $w_j$ is the weight assigned to the positive samples in the $j$-th image pair, and $n_{neg}^{j}$ and $n_{pos}^{j}$ are the number of negative and positive samples, respectively. In the $w_j$, $\alpha$ and $\beta$ control the degree to which the weight varies with the ratio of positive-negative samples, and $\lambda$ controls the basic weight value when this ratio is large enough. More specifically, $w_j$ is inversely correlated to the ratio of positive-negative samples, as shown in Fig. \ref{fig14}. The hyper-parameters $\alpha$, $\beta$, and $\lambda$ are empirically set to 2, 0.002, and 1.5, respectively.

\subsection{Domain Difference Removal Module} \label{c}
The domain shifts between bitemporal images are an urgent and rarely attended issue in the current change detection task, which interferes with recognizing truly changed pixels by the network. To enable the model to focus on domain-agnostic features, eliminating domain-specific style in the features of bitemporal images is the key. Therefore, based on the claim in the introduction that the channel-wise mean and standard deviation of local-level features can proxy for style, we design the domain difference removal (DDR) module and its variant to provide more possibilities for the decorrelation of features, which are described below.

\subsubsection{Global-to-Local Normalization}
Our motivation comes from the existing observation that the styles mainly manifest in the shallow layers of convolutional neural networks (CNNs) while the contents mainly lie in the higher layers and partially shallow layers. Since instance normalization (IN) is usually combined with CNNs to assist in removing feature variance caused by style shifts \cite{nam2018batch,xie2017aggregated,pan2018two}, we introduce local IN (LIN) layer to the first two residual blocks in SD network following the above observation. In this way, the domain characteristics of bitemporal features are reasonably reduced before calculating the difference between them. For the feature matrix $\textbf f_j \in \mathbb{R}^{C \times H \times W} (j \in \{1,2,\dots,N\})$ of an image of the $j$-th image pair in the mini-batch, we divide it into local regions and the features in $r$-th rectangular region are represented as $\textbf f_{jr} \in \mathbb{R}^{C \times \frac{H}{\lambda} \times \frac{W}{\lambda}}$, where $C$, $H$, and $W$ represent the channel number, height, and width of the feature map.
$\tilde f_{jr}^{chw}$ denotes the original element at position $(h,w)$ in $c$-th channel of $\textbf f_{jr}$, then its LIN process can be given by,
\begin{equation} 
\begin{aligned}
\bm\mu_{lin}^{jc}&=\frac{\lambda_{}^{2}}{HW}\sum_{w=1}^{\frac{W}{\lambda}}\sum_{h=1}^{\frac{H}{\lambda}}\tilde f_{jr}^{chw},\\    
\bm\Sigma_{lin}^{jc}&=\frac{\lambda_{}^{2}}{HW}\sum_{w=1}^{\frac{W}{\lambda}}\sum_{h=1}^{\frac{H}{\lambda}}(\tilde f_{jr}^{chw}-\bm\mu_{lin}^{jc})_{}^{2}+\epsilon,\\
f_{jr}^{chw}&={\bm\Sigma_{lin}^{jc}}_{}^{-\frac{1}{2}}(\tilde f_{jr}^{chw}-\bm\mu_{lin}^{jc}),  \label{eq2}
\end{aligned}
\end{equation}
where $\bm\mu_{lin}^{jc}$ and $\bm\Sigma_{lin}^{jc}$ are the mean and variance calculated from the $c$-th channel of $\textbf f_{jr}$, respectively, and $\epsilon > 0$ is a small positive number for preventing a singular $\bm\Sigma$.
However, the RIN layers reduce the discriminability of instances, but the BN layers are beneficial to preserve this discriminative information. Thus, to trade off domain invariance and feature discriminability and benefit from each other, the RIN layer is added after each residual block's last ReLU activation function while retaining the original two BN layers.
Fig. \ref{fig3}(b) shows the structure of the residual block embedded with the \emph{global-to-local normalization}. Let $\textbf f \in \mathbb{R}^{C \times N\times H\times W}$ denotes the feature matrix of a mini-batch, $\tilde f_{j}^{chw}$ denotes the original element at position $(h,w)$ in $c$-th channel of $\textbf f_{j}$, the BN can be calculated,
\begin{equation} 
\begin{aligned}
\bm\mu_{bn}^{c}&=\frac{1}{NHW}\sum_{j=1}^{N}\sum_{w=1}^{W}\sum_{h=1}^{H}\tilde f_{j}^{chw},\\    
\bm\Sigma_{bn}^{c}&=\frac{1}{NHW}\sum_{j=1}^{N}\sum_{w=1}^{W}\sum_{h=1}^{H}(\tilde f_{j}^{chw}-\bm\mu_{bn}^{c})_{}^{2}+\epsilon,\\
f_{j}^{chw}&={\bm\Sigma_{bn}^{c}}_{}^{-\frac{1}{2}}(\tilde f_{j}^{chw}-\bm\mu_{bn}^{c}).  \label{eq3}
\end{aligned}
\end{equation}
where $\bm\mu_{bn}^{c}$ and $\bm\Sigma_{bn}^{c}$ are the mean and variance calculated from the $c$-th channel of $\textbf f_{j}$, respectively.
Then, the encoded features $\textbf f_{j\_gln} \in \mathbb{R}^{C \times H\times W}$ through the \emph{global-to-local normalization (gln)} are denoted as,
\begin{equation} 
\textbf f_{j\_gln}=[LIN(BN(\textbf f_{j}))].  \label{eq4}
\end{equation}
Both BN and LIN benefit from each other such that CNNs preserves the discriminability of individual samples while being less vulnerable to appearance changes. 
\subsubsection{Global-to-Local Whitening}
These normalization methods (BN, LIN) center and scale the activations (features) across different dimensions to standardize the distribution. Nevertheless, these activations are not decorrelated, so the correlation between them remains. This problem leads to suboptimal optimization efficiency of the network. As a promising data preprocessing approach, whitening transformation \cite{mohamed1998simple} is a classic technique to remove the correlation between features. Previous works \cite{pan2019switchable,huang2018decorrelated,cho2019image} have demonstrated that this technique preserves the desirable properties of normalization and further improves normalization's optimization efficiency and the generalization ability of feature representations. 
Therefore, we further propose an enhanced version of \emph{global-to-local normalization}, namely \emph{global-to-local whitening}. The normalization operations in LIN and BN layers are replaced with whitening to obtain local instance whitening (LIW) layers and batch whitening (BW) layers, as shown in Fig. \ref{fig3} (c). Taking the feature matrix $\textbf f$ as an example, the whitening transformation $\Phi (\cdot): \mathbb{R}^{C \times HW}\rightarrow \mathbb{R}^{C \times HW}$ can be formulated,
\begin{equation} 
\begin{aligned}
\Phi (\textbf f)={\bm\Sigma_{bw}}_{}^{-\frac{1}{2}}(\textbf f-\bm\mu_{bw} \cdot \bm1_{}^{T}), \label{eq5}
\end{aligned}
\end{equation}
note that where $\bm\Sigma_{bw}$ is the calculated covariance matrix, and $\bm1 \in \mathbb{R}^{HW}$ is a column vector of all ones. The whitening transformation ensures that for the transformed feature matrix $\Phi (\textbf f)$ is whitened, that is, $\Phi (\textbf f)$ and the original feature matrix $\textbf f$ satisfies that $\Phi (\textbf f) \cdot (\Phi (\textbf f))_{}^{T}=(HW) \cdot \textbf I \in \mathbb{R}^{C \times C}$, where $\textbf I$ represents the identity matrix. Also, different whitening methods can be implemented by calculating $\bm\mu_{bw}$ and $\bm\Sigma_{bw}$ values for different sets of features. We describe in detail how they are calculated below.

\textbf{Local Instance Whitening (LIW)}.
In LIW, whitening is performed on local regions of a single sample, and $\bm\mu_{liw}$ and $\bm\Sigma_{liw}$ are computed within the feature matrix of each instance. Given the feature matrix $\textbf f_{jr}$ of the $r$-th region in a single sample, $\bm\mu_{liw}$ and $\bm\Sigma_{liw}$ are given by,
\begin{gather} 
\bm\mu_{liw}=\frac{\lambda_{}^{2}}{HW}\textbf f_{jr} \cdot \bm1,\notag \\   
\bm\Sigma_{liw}=\frac{\lambda_{}^{2}}{HW}(\textbf f_{jr}-\bm\mu_{liw} \cdot \bm1_{}^{T})(\textbf f_{jr}-\bm\mu_{liw} \cdot \bm1_{}^{T})_{}^{T}+\epsilon\textbf I.  \label{eq6}
\end{gather}
In this way, the whitening transformation $\Phi (\cdot)$ whitens each single sample separately.

\textbf{Batch Whitening (BW)}. 
In contrast, BW \cite{huang2018decorrelated} applies the whitening to a whole batch of sample instead of single ones. For the feature matrix $\textbf f$ of a mini-batch, the two statistics in whitening are calculated as follows,
\begin{gather}
\bm\mu_{bw}=\frac{1}{NHW}\textbf f \cdot \bm1,\notag \\    
\bm\Sigma_{bw}=\frac{1}{NHW}(\textbf f-\bm\mu_{bw} \cdot \bm1_{}^{T})(\textbf f-\bm\mu_{bw} \cdot \bm1_{}^{T})_{}^{T}+\epsilon\textbf I,  \label{eq7}
\end{gather}
then BW whitens all samples in a mini-batch.

Next, the calculated covariance matrix $\bm\Sigma_{liw/bw}$ can be eigen decomposed into $\textbf Q \Lambda \textbf Q_{}^{T}$, so we can calculate the inverse square root of $\bm\Sigma_{liw/bw}$ by the following formula,
\begin{equation} 
\begin{aligned}
{\bm\Sigma_{liw/bw}}_{}^{-\frac{1}{2}}=\textbf Q \Lambda_{}^{-\frac{1}{2}} \textbf Q_{}^{T},   \label{eq8}
\end{aligned}
\end{equation}
where $\textbf Q$ is an orthogonal matrix composed of the eigenvectors of $\bm\Sigma_{liw/bw}$, $\Lambda$ is a diagonal matrix and $\Lambda_{ii}$ corresponds to the eigenvalue of the eigenvector of the $i$-th column in matrix $\textbf Q$. Also, to alleviate the expensive computation of eigenvalue decomposition, we borrow the Newton's iterations in \cite{huang2019iterative} to approximate the whitening matrix.

The features $\textbf f_{j\_glw} \in \mathbb{R}^{C \times H\times W}$ through the \emph{global-to-local whitening (glw)} are then encoded as,
\begin{equation}
\begin{aligned}
\textbf f_{j\_glw}=[LIW(BW(\textbf f_{j}))].  \label{eq9}
\end{aligned}
\end{equation}

Note that in the whitening transformation, the diagonal elements of the covariance matrix $\bm\Sigma_{liw/bw}$ are the variances of each channel, and the off-diagonal elements are the correlations between channels. However, the $\bm\Sigma_{lin/bn}$ calculated in the normalization is the variance matrix by zeroing the off-diagonal elements, which simply takes into account the correlations of the same channels. Thus, compared to \emph{global-to-local normalization}, our designed variant \emph{global-to-local whitening} based on whitening provides better optimization conditions or style invariance for encoding features, giving more possibilities for decorrelation of features. 

\begin{figure}[t] 
    \setlength{\abovecaptionskip}{-0.35cm}
    \begin{center}
    \centering 
    \includegraphics[width=0.4\textwidth, height=0.28\textwidth]{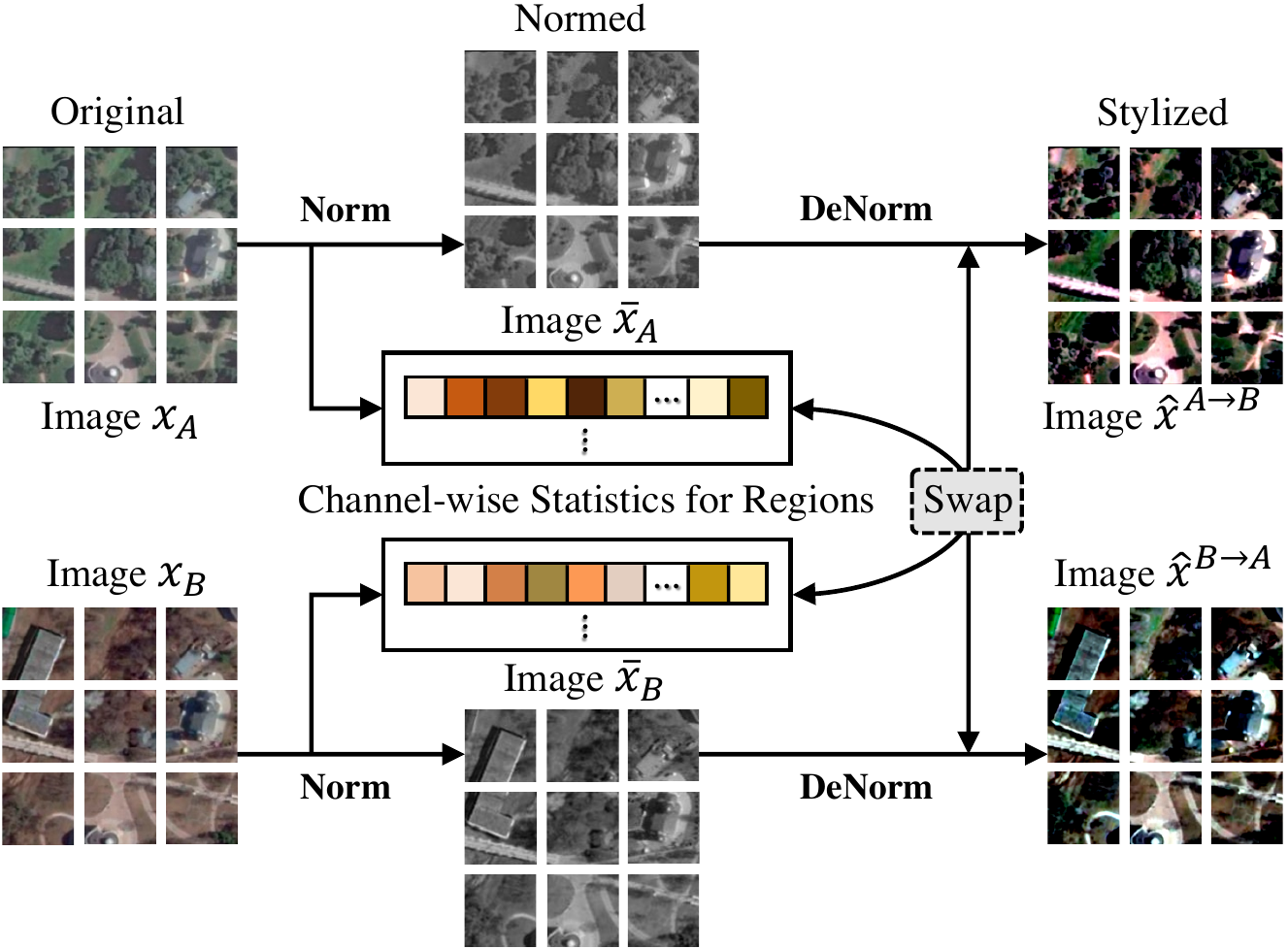} 
    \end{center}
    \caption{Demonstration of the style transformation strategy in the cross-temporal style transformation (CTST) module. Channel-wise statistics are swapped in the region-level normalized images to generate a stylized image pair ($\hat x_{}^{A\to B}$, $\hat x_{}^{B\to A}$).} 
    \label{fig5}
\end{figure}

\subsection{Cross-Temporal Generalization Learning} \label{d}
\subsubsection{Cross-Temporal Style Transformation Module}
\begin{figure}[t] 
    \setlength{\abovecaptionskip}{-0.35cm}
    \begin{center}
    \centering 
    \includegraphics[width=0.45\textwidth, height=0.28\textwidth]{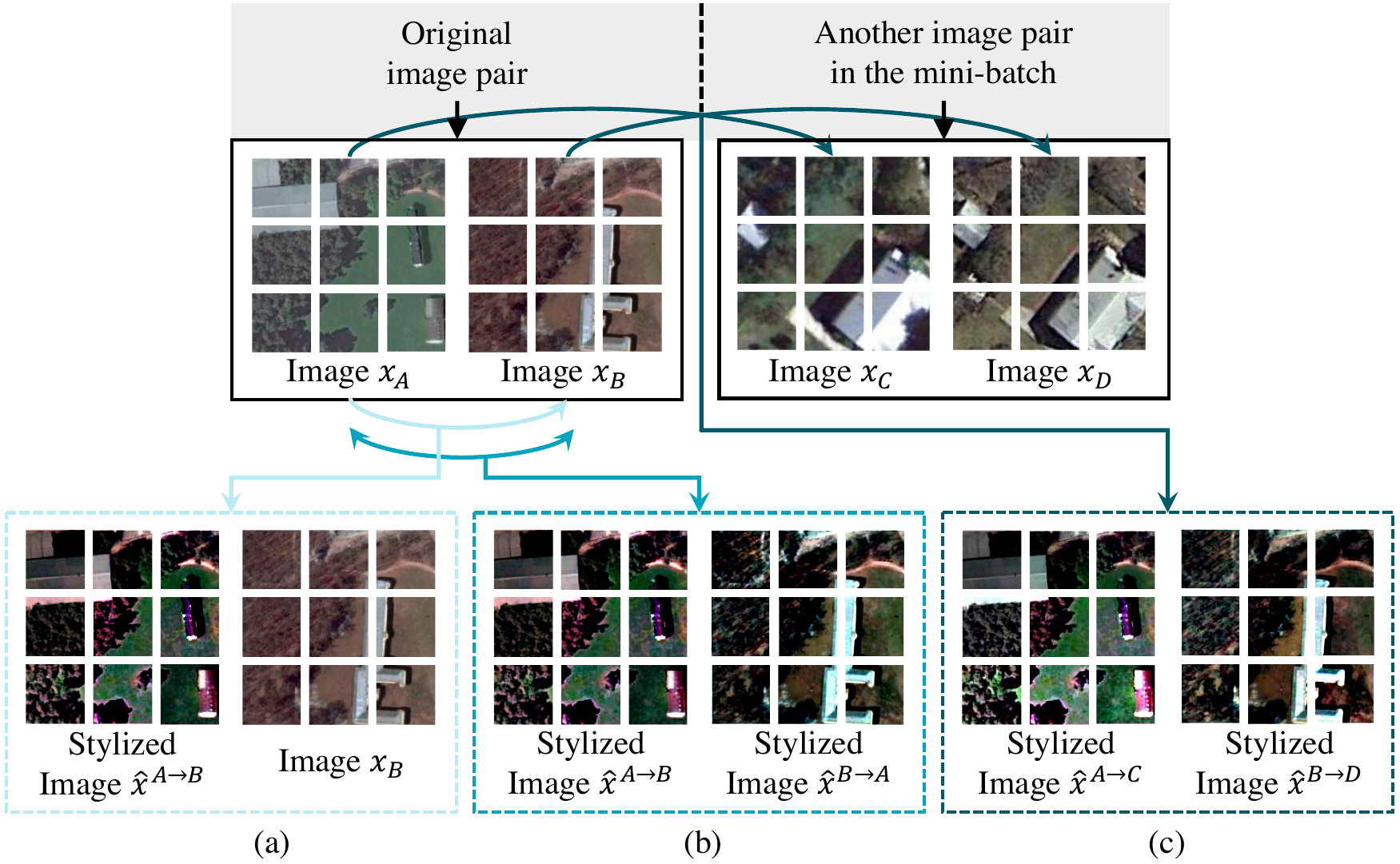} 
    \end{center}
    \caption{Three transformation modes in the cross-temporal style transformation (CTST) module: (a) Unilateral style transformation (UST). (b) Bilateral style transformation (BST). (c) Intra-batch style transformation (IBST).} 
    \label{fig6}
\end{figure}

We assume that highlighting images' class (object) characteristics can assist the model in learning feature representations robust to the domain shift. Therefore, to further avoid the overdetection of pseudo-changed pixels, we propose performing the cross-temporal style transformation (CTST) on random samples in bitemporal image pairs. CTST aims to avoid the network from overfitting the low-order statistics that carry the style information, thus focusing more on the image content when deciding whether pixels are changed.
Based on the claim in the introduction that the channel-wise mean and standard deviation of local-level images can proxy for style, style transformation is achieved by updating both statistics of the image within a bitemporal pair. Given images $x_A$ and $x_B$ from different phases, we design three different style transformation modes: unilateral style transformation (UST), bilateral style transformation (BST), and intra-batch style transformation (IBST). Fig. \ref{fig5} is a schematic example of the CTST module. Three modes are specifically shown in Fig. \ref{fig6}.

\textbf{Normalized Local Image.} For arbitrary image $x_j\in \mathbb{R}^{H' \times W'}$ of the $j$-th image pair in the mini-batch, we first divide it into local regions and the pixels in $r$-th rectangular region are represented as $x_{jr} \in \mathbb{R}^{3 \times \frac{H'}{\lambda'} \times \frac{W'}{\lambda'}}$. We then compute the channel-wise mean $\bm\mu_{rst}$ and variance $\bm\Sigma_{rst}$ of $x_{jr}$ and obtain the normalized image $\bar x_{jr}$ by normalization, 
\begin{equation}
\begin{aligned}
\bm\mu_{rst}^{jc}&=\frac{{\lambda'}_{}^{2}}{H'W'}\sum_{w=1}^{\frac{W'}{\lambda}}\sum_{h=1}^{\frac{H'}{\lambda}}x_{j}^{chw},\\
\bm\Sigma_{rst}^{jc}&=\frac{{\lambda'}_{}^{2}}{H'W'}\sum_{w=1}^{\frac{W'}{\lambda}}\sum_{h=1}^{\frac{H'}{\lambda}}(x_{j}^{chw}-\bm\mu_{rst}^{jc})_{}^{2}+\epsilon,\\
{\bar x}_{j}^{chw}&={\bm\Sigma_{rst}^{jc}}_{}^{-\frac{1}{2}}(x_{j}^{chw}-\bm\mu_{rst}^{jc}),  \label{eq17}
\end{aligned}
\end{equation}
where $x_{j}^{chw}$ denotes the original element at position $(h,w)$ in $c$-th channel of $x_{jr}$. Next, the style transformation of the three modes can be performed based on the normalized image.

\begin{itemize}
\item \textbf{Unilateral Style Transformation:}
\end{itemize}

In the UST mode, we only transform the style of $x_A$ or $x_B$ in the original image pair to obtain a stylized image $\hat x_{}^{A\to B}$ or $\hat x_{}^{B\to A}$. Taking $\hat x_{}^{A\to B}$ as an example, the mean $\bm\mu_{rst}^{\scriptscriptstyle Bjc}$ and variance $\bm\Sigma_{rst}^{\scriptscriptstyle Bjc}$ of local-level image $x_{\scriptscriptstyle Br}$ are used as style information to assign to the local-level normalized image $\bar x_{\scriptscriptstyle Ar}$. The stylized image is generated by reconstruction with $\bm\mu_{rst}^{\scriptscriptstyle Bjc}$, $\bm\Sigma_{rst}^{\scriptscriptstyle Bjc}$, and $\bar x_{\scriptscriptstyle Ar}$,
\begin{equation} 
\hat x_{}^{A\to B}=\bar x_{\scriptscriptstyle Ar}{\bm\Sigma_{rst}^{\scriptscriptstyle Bjc}}_{}^{-\frac{1}{2}}+\bm\mu_{rst}^{\scriptscriptstyle Bjc}. \label{eq18}
\end{equation}
$\hat x_{}^{B\to A}$ is generated in a similar principle to $\hat x_{}^{A\to B}$.
Then, a new image pair $(\hat x_{}^{A\to B}, x_B)$ or $(x_A, \hat x_{}^{B\to A})$ is produced by the UST mode and used for training the SD network.

\begin{itemize}
\item \textbf{Bilateral Style Transformation:}
\end{itemize}

In the BST mode, we perform style transformation on both $x_A$ and $x_B$ in bitemporal image pair to generate stylized images $\hat x_{}^{A\to B}$ and $\hat x_{}^{B\to A}$, and the two stylized images are obtained with Eq. (\ref{eq17}) and Eq. (\ref{eq18}). Thereafter, a new image pair is combined and added to the training data, \emph{i.e.}, $(\hat x_{}^{A\to B}, \hat x_{}^{B\to A})$.

\begin{itemize}
\item \textbf{Intra-Batch Style Transformation:}
\end{itemize}

For the third mode IBST, in addition to styles within a given bitemporal image pair, we consider introducing diverse style information from other image pairs in the mini-batch. Given another image pair $(x_C, x_D)$, we swap the both statistics of $x_C$ and $x_D$ with those of $x_A$ and $x_B$, respectively, producing stylized images $\hat x_{}^{A\to C}$ and $\hat x_{}^{B\to D}$. Furthermore, the formula calculation is similar to Eq. (\ref{eq17}) and Eq. (\ref{eq18}), and a new image pair is denoted as $(\hat x_{}^{A\to C}, \hat x_{}^{B\to D})$.

Finally, three style transformations are randomly selected following a uniform distribution for the learning of the SD network, which enables the network to achieve the best detection performance (Verified in Table \ref{table5}).

\subsubsection{Cross-Temporal Consistency Regularization Loss} \label{4.4.2}
The above cross-temporal style transformation implies a strong implicit constraint that requires pixels before and after stylization to be predicted as the same class.
However, due to the transformation of style, the knowledge learned by the network from the original image pair may conflict with that learned from the newly generated images. To alleviate this disagreement, we add an explicit cross-temporal consistency regularization (CTCR) loss $\mathcal L_{CTCR}$ to align the predictions ($\textbf P_{out}^{sty}$, $\textbf P_{out}^{ori}$) of the stylized and original image pairs in the output space,
\begin{equation}
\begin{aligned}
\mathcal L_{CTCR}=\textbf {KL}(\textbf P_{out}^{sty}, \textbf P_{out}^{ori}).  \label{eq19}
\end{aligned}
\end{equation}
$\textbf {KL} (\cdot)$ denotes the calculation of Kullback-Leibler divergence. Such a constraining process forces the network to focus on invariant content information, thus achieving knowledge consistency among bitemporal image pairs of various views.

\section{Experiments}
\subsection{Datasets and Experimental Setup}
\subsubsection{Datasets}  \label{4.1}
To validate the generalization of our method, we conduct experiments on five public datasets, including remote sensing and natural datasets.

\noindent\textbf{Remote sensing datasets:}

\textbf{SVCD} \cite{lebedev2018change} consists of real season-varying images. There are a total of 11 pairs of season-varying images obtained by Google Earth, including 7 pairs with the original size of $4725 \times 2700$ and 4 pairs with the original size of $1900 \times 1000$. The resolution of the images is from 3 cm to 100 cm per pixel. All images are further cropped to $256 \times 256$ image pairs. We use 10000 pairs as the training set, 3000 pairs as the validation set, and 3000 pairs as the testing set. The ground-truth labels for this dataset do not account for pixels that cause differences between image pairs due to seasonal factors. This is undoubtedly a great challenge compared to other datasets.

\begin{table}[t]  
  \centering
  \caption{Details of publicly available datasets before preprocessing.}
    \begin{tabular}{cccc}
    \toprule
   Dataset &Image Pairs & Image Size & Resolution \\
    \midrule
    \multirow{2}[0]{*}{SVCD} &7 & $4725 \times 2700$ & 0.03 \\
      &4 & $1900 \times 1000$ & 1.0 \\
    \midrule
    SZADA &7 &$952 \times 640$ & 1.5 \\
    SYSU-CD &20000 & $256 \times 256$&0.5  \\
    PCD &200 & $224 \times 1024$&-  \\
    VL-CMU-CD &1362 & $1024 \times 768$&-  \\
    \bottomrule
    \end{tabular}
  \label{add}
\end{table}

\textbf{SZTAKI} \cite{benedek2009change} consists of three subsets (SZADA, TISZADOB, and ARCHIVE) split by region and is provided by the Hungarian Institute of Geodesy Cartography \& Remote Sensing (F$\rm\ddot{O}$MI) and Google Earth. We only use the SZADA dataset in the SZTAKI, consisting of 7 optical aerial image pairs at the resolution of 1.5 m per pixel and covering in aggregate 9.5 $\rm km_{}^{2}$ area. The size of each image in this dataset is $952 \times 640$. Six image pairs are used for training and the top left $752 \times 448$ rectangle of one image pair is used for testing. We further crop them into $256 \times 256$ pairs, including 72 training pairs and 6 testing pairs. There are many types of changes, such as roads or farmland. Note that the number of image pairs in the SZADA dataset is much less than in the other two datasets, which easily leads to network overfitting and examines the stability of all methods.

\textbf{SYSU-CD} \cite{shi2021deeply} is provided by Sun Yat-Sen University and is captured in a populous metropolis in southern China, Hong Kong, covering a total land area of 1106.66 $\rm km_{}^{2}$. This dataset contains 20,000 aerial image pairs taken in 2007 and 2014, respectively. Each image has a size of $256 \times 256$ and a resolution of 0.5 m per pixel. Following \cite{shi2021deeply}, we divide the original 20,000 image pairs into 12,000 training pairs, 4,000 validation pairs, and 4,000 test pairs. In addition to improvements in dataset volume and image resolution, this dataset greatly complements change samples of high-rise buildings and port-related compared to previous datasets. 

\noindent\textbf{Natural datasets:}

\textbf{PCD} \cite{jst2015change} is divided into two subsets, ``GSV" and ``TSUNAMI", each comprising 100 pairs of panoramic images with corresponding hand-drawn change masks. The images within each pair are captured from varying camera angles. The GSV subset focuses on Google Street View imagery, while the TSUNAMI subset features post-tsunami scenes, both providing 100 panoramic image pairs for analysis.

\textbf{VL-CMU-CD} \cite{alcantarilla2018street} is designed for street-view change detection over an extended time period. It includes 151 sequences of images, resulting in 1,362 image pairs available for analysis. Each image pair is accompanied by a labeled mask representing five distinct classes as ground truth. The dataset is divided into 933 image pairs for training and 429 image pairs for testing.

\subsubsection{Implementation Details} The proposed algorithm is implemented in PyTorch 1.8 on an NVIDIA RTX 3090 GPU with 24GB of memory. The batch size is set to 8. The stochastic gradient descent (SGD) is adopted as our optimizer to optimize the network parameters, where the weight decay rate is set to $ 1 \times 10_{}^{-8} $ and momentum is set to 0.9. The initial learning rate is set to $ 1 \times 10_{}^{-2} $, which is decayed following a polynomial learning rate scheduling with a power of 0.9 during training.

\begin{table}[t] 
  \centering
  \caption{Comparison results of all methods on the SVCD dataset.}
  \resizebox{\linewidth}{!}{
    \begin{tabular}{c|ccccc}
    \toprule
    Method & Pre.(\%) & Rec.(\%) & F1.(\%) & IoU(\%) & OA(\%) \\
    \midrule
    FC-EF \cite{daudt2018fully} &77.48 &88.13 &82.46 &71.15&92.29  \\
    FC-Siam-conc \cite{daudt2018fully} &72.88 &89.21 &80.22 &67.98 &93.10  \\
    FC-Siam-diff-res \cite{daudt2018fully} &82.87 &93.35 &87.80 &79.25 &93.29  \\
    FCN-PP \cite{lei2019landslide} &83.05 &91.80 &87.21 &77.31 &94.19  \\
    W-Net \cite{hou2019w} &85.25 &92.90 &88.91 &80.04 &95.21  \\
    CDGAN \cite{hou2019w} &89.05 &89.47 &89.26 &80.60 &95.39  \\
    STANet \cite{chen2020spatial} &89.24 &93.51 &91.33 &84.04 &95.99  \\
    DSAMNet \cite{shi2021deeply} &93.35 &92.41 &92.90 &86.60 &96.31  \\
    ESCNet \cite{zhang2021escnet} &90.03 &\textbf{95.82} &92.83 &86.63 &97.29 \\
    SEIFNet \cite{huang2024spatiotemporal} &92.15 &93.07 &92.61 &86.23 &96.11 \\
    MFINet \cite{ren2024mfinet} &93.10 &94.51 &93.80 &88.32 &96.53 \\
    SAAN \cite{guo2024saan} &95.91 &95.63 &95.77 &91.88 &97.98 \\
    DAMFANet \cite{ren2024dual} &\underline{96.20} &95.50 &\underline{95.85} &\underline{92.03} &\underline{98.50} \\
    \midrule
    DonaNet (Ours) &\textbf{96.93} &\underline{95.80}  &\textbf{96.36} &\textbf{92.98} &\textbf{99.10} \\
    \bottomrule
    \end{tabular}}
  \label{table1}
\end{table}

To evaluate the performance of our method, we utilize six standard evaluation metrics, precision, recall, F1-score, intersection over union (IoU), and overall accuracy (OA), which are given by,
\begin{equation}
\text{Precision}=\frac{\text{TP}}{\text{TP}+\text{FP}}.
\end{equation} 
\begin{equation} 
\text{Recall}=\frac{\text{TP}}{\text{TP}+\text{FN}}.
\end{equation}
\begin{equation}
\text{F1-score}=\frac{\text{2} \times \text{precision} \times \text{recall}}{\text{precision}+\text{recall}}.
\end{equation} 
\begin{equation}
\text{IoU}=\frac{\text{Detection Result}\cap \text{Ground Truth}}{\text{Detection Result}\cup \text{Ground Truth}}.
\end{equation}
\begin{equation} 
\text{OA}=\frac{\text{TP}+\text{TN}}{\text{TP}+\text{TN}+\text{FP}+\text{FN}}.
\end{equation}
TP, TN, FP, and FN represent the sample numbers of true positives, true negatives, false positives, and false negatives. PC indicates the true consistency that equals to OA.

\begin{figure}[t]  
    \setlength{\abovecaptionskip}{-0.35cm}
    \begin{center}
    \centering 
    \includegraphics[width=0.36\textwidth, height=0.32\textwidth]{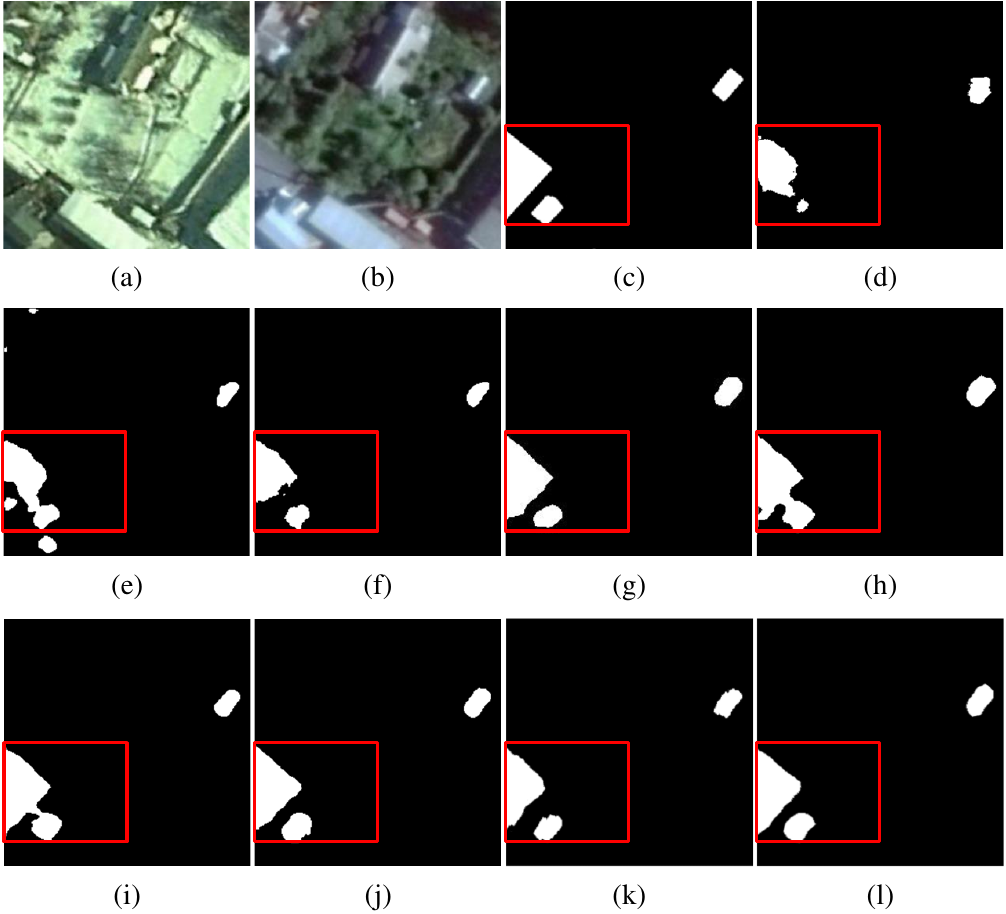} 
    \end{center}
    \caption{Quantitative comparison results of different methods on the SVCD dataset. (a) Image $x_A$. (b) Image $x_B$. (c) Ground truth. (d) FC-EF. (e) FC-Siam-conc. (f) FC-Siam-diff-res. (g) STANet. (h) DSAMNet. (i) ESCNet. (j) DonaNet-base. (k) DonaNet-glw. (l) DonaNet.} 
    \label{fig7}    
\end{figure}

\subsection{Comparison with Existing Methods}
\subsubsection{Comparison on the SVCD Dataset} 
\begin{figure}[t]  
    \setlength{\abovecaptionskip}{-0.35cm}
    \begin{center}
    \centering 
    \includegraphics[width=0.36\textwidth, height=0.32\textwidth]{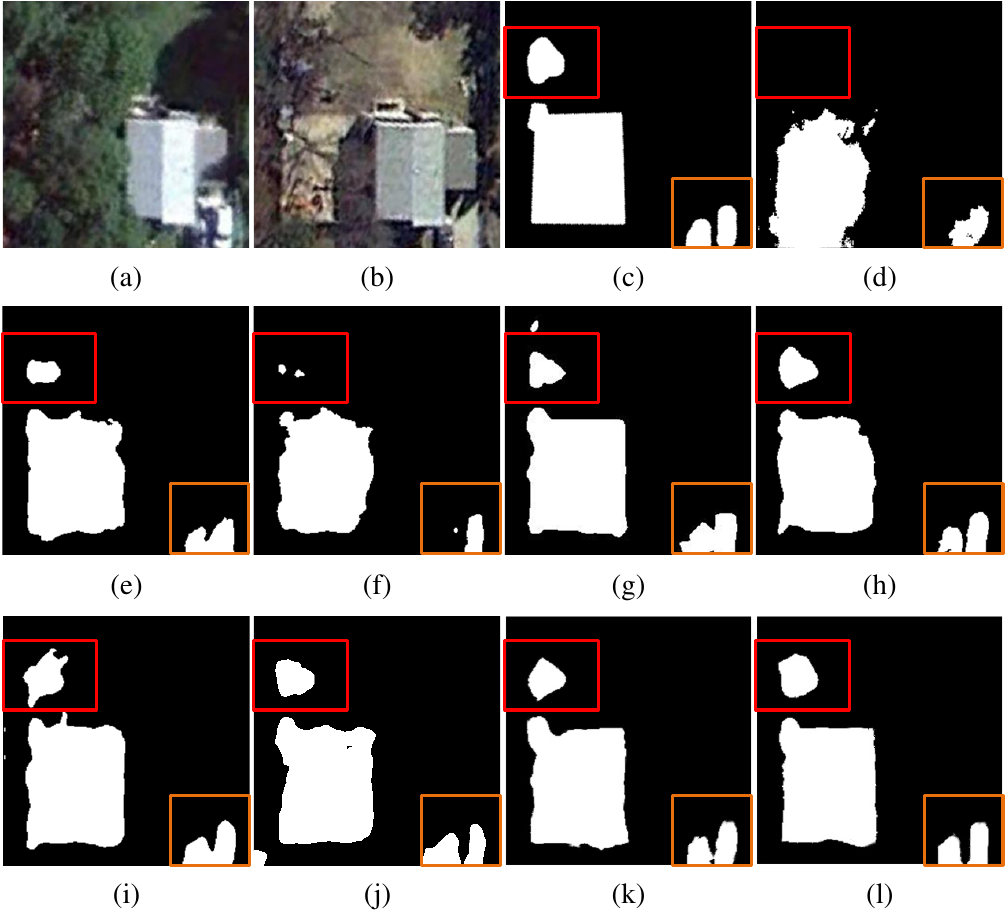} 
    \end{center}
    \caption{Quantitative comparison results of different methods on the SVCD dataset. (a) Image $x_A$. (b) Image $x_B$. (c) Ground truth. (d) FC-EF. (e) FC-Siam-conc. (f) FC-Siam-diff-res. (g) STANet. (h) DSAMNet. (i) ESCNet. (j) DonaNet-base. (k) DonaNet-glw. (l) DonaNet.} 
    \label{fig8}    
\end{figure}

$\bullet$\ \textbf{Qualitative Comparison.} The visual comparisons of experimental results on the SVCD dataset are demonstrated in Fig. \ref{fig7} and Fig. \ref{fig8}. As shown in Fig. \ref{fig7}(d)-(i) and Fig. \ref{fig8}(d)-(i), the boundaries of detection results from nine existing methods are ambiguous and non-smooth. Among these methods, false detections often occur, and the contours of changed objects can only be detected roughly. For example, the vertical edges of the building in Fig. \ref{fig7}(d)-(i) are not clearly detected but are curved. In Fig. \ref{fig8}(d)-(i), there is a similar false detection situation for the detected edge of the rectangular land. Although DonaNet-glw's fitting of object edges is slightly worse, by further adding the CTGL module, DonaNet almost perfectly alleviates the edge ambiguity problem. Besides, qualitative comparisons for pseudo-changes are analyzed in detail,
\noindent\textbf{Analysis of Pseudo-changes:} Due to style differences caused by variations in illumination and shadows, the existing methods produce overdetection for those pixels with style shifts. From the red rectangular boxes in Fig. \ref{fig7}(h)-(i) and Fig. \ref{fig8}(g)-(i), we can see that the models STANet, DSAMNet, and ESCNet incorrectly detected roads as changed pixels while these roads only appear in different colors in two phases, which significantly affects the accuracy of detection results. In contrast, our DonaNet-glw and DonaNet achieve the most superior detection results and avoids the overdetection for pseudo-changed pixels. This indicates that DonaNet has a better trade-off between feature discriminability and robustness to domain shifts. 

\begin{table}[t] 
  \centering
  \caption{Comparison results of all methods on the SZADA dataset.}
  \resizebox{\linewidth}{!}{
    \begin{tabular}{c|ccccc}
    \toprule
    Method & Pre.(\%) & Rec.(\%) & F1.(\%) & IoU(\%)& OA(\%)  \\
    \midrule
    FC-EF \cite{daudt2018fully} &46.41 &51.71 &48.92 &32.38&80.47 \\
    FC-Siam-conc \cite{daudt2018fully} &44.97 &44.17 &44.57 &28.67&83.14  \\
    FC-Siam-diff-res \cite{daudt2018fully} &47.03 &48.64 &47.82 &31.42&84.94  \\
    FCN-PP \cite{lei2019landslide} &44.74 &\textbf{54.03} &48.95 &32.40&85.47  \\
    W-Net \cite{hou2019w} &46.11 &45.01 &45.55 &29.49&86.72  \\
    CDGAN \cite{hou2019w} &44.90 &46.91 &45.88 &29.77&85.92  \\
    STANet \cite{chen2020spatial} &50.04 &51.71 &50.86 &34.10&89.85  \\
    DSAMNet \cite{shi2021deeply} &61.70 &41.14 &49.36 &32.77&91.59  \\
    ESCNet \cite{zhang2021escnet} &49.38 &52.06 &50.68 &33.94 &90.37  \\
    SEIFNet \cite{huang2024spatiotemporal} &58.04 &46.10 &51.39 &34.58 &88.54 \\
    MFINet \cite{ren2024mfinet} &59.83 &48.15 &53.36 &36.39 &90.61 \\
    SAAN \cite{guo2024saan} &61.57 &50.26 &55.34 &38.26 &92.04 \\
    DAMFANet \cite{ren2024dual} &\underline{62.11} &51.02 &\underline{56.02} &\underline{38.91} &\underline{93.13} \\
    \midrule
    DonaNet (Ours) &\textbf{64.86} &\underline{52.51} &\textbf{58.04} &\textbf{40.88}&\textbf{95.70}  \\
    \bottomrule
    \end{tabular}}
  \label{table2} 
\end{table}

\begin{figure*}[t] 
    \setlength{\abovecaptionskip}{-0.35cm}
    \begin{center}
    \centering 
    \includegraphics[width=0.75\textwidth, height=0.18\textwidth]{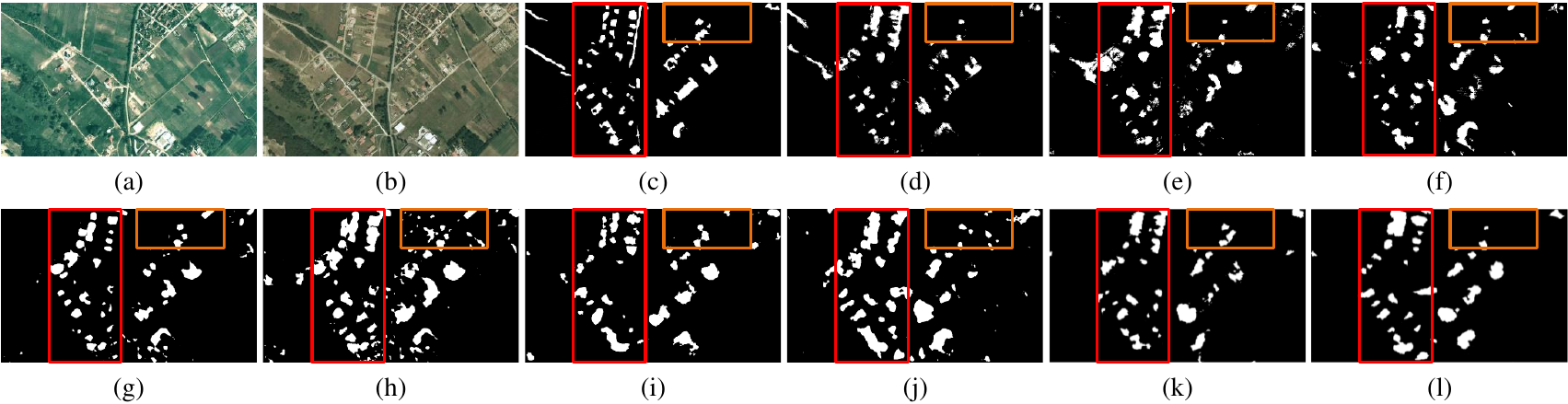} 
    \end{center}
    \caption{Quantitative comparison results of different methods on the SZADA dataset. (a) Image $x_A$. (b) Image $x_B$. (c) Ground truth. (d) FC-EF. (e) FC-Siam-conc. (f) FC-Siam-diff-res. (g) STANet. (h) DSAMNet. (i) ESCNet. (j) DonaNet-base. (k) DonaNet-glw. (l) DonaNet.} 
    \label{fig10}    
\end{figure*}

$\bullet$\ \textbf{Quantitative Comparison.} To further illustrate the performance of our method, the quantitative evaluation results are reported in Table \ref{table1}. As shown in Table \ref{table1}, DSAMNet achieves the second-highest precision and F1-score, but its recall is not very ideal, indicating that there are many undetected pixels. ESCNet obtains the highest recall while its precision is the third highest, so we infer that the overdetection phenomenon of this method should be relatively serious. STANet also shows decent performance on the recall, F1-score, and OA. Among these methods, our method achieves the best Precision, F1-score, IoU, and OA, surpassing the second-highest results of these metrics by $0.73\%$, $0.51\%$, $0.95\%$, and $0.60\%$. These quantitative comparison results further illustrate the effectiveness of our method.

\subsubsection{Comparison on the SZADA Dataset}
$\bullet$\ \textbf{Qualitative Comparison.} Fig. \ref{fig10} demonstrates the visual comparison results on the SZADA dataset. From Fig. \ref{fig10}(a)-(c), we can see that the changed regions in this dataset are small-sized and sparse, which brings challenges to the detection task. Many over-detected pixels occur in the detection results obtained by the models FC-EF, FC-Siam-conc, and FC-Siam-diff-res, and their visual effects are relatively unclean, \emph{i.e.}, the ``salt-and-pepper" phenomenon, as shown in Fig. \ref{fig10}(d)-(f). In Fig. \ref{fig10}(g)-(i), the detection results of the FCN-PP, W-Net, and CDGAN methods do not have spots but not refined, \emph{e.g.}, they merge the outlines of multiple small buildings into one. The visual result of ESCNet is better than other existing methods, but the detected edges are discontinuous. Although our method DonaNet-glw and DonaNet suffer from some missed detections, they achieve more accurate and refined detection results through stronger feature representation capabilities. Besides, qualitative comparisons for pseudo-changes are analyzed in detail,
\noindent\textbf{Analysis of Pseudo-changes:} Due to style differences caused by variations in illumination and seasons, overdetection of pseudo-changed pixels still occurs in existing methods. In the area corresponding to the orange rectangular box in Fig. \ref{fig10}, the same cropland presents a completely different appearance in different phases. As shown in Fig. \ref{fig10}(e)-(i), most methods mistakenly detect these unchanged cropland as changed areas. In contrast, by adding cross-temporal generalization learning to DonaNet-glw, DonaNet effectively overcomes the problem of pseudo-changes and achieves the most accurate detection results.

\begin{figure}[t] 
    \setlength{\abovecaptionskip}{-0.35cm}
    \begin{center}
    \centering 
    \includegraphics[width=0.36\textwidth, height=0.32\textwidth]{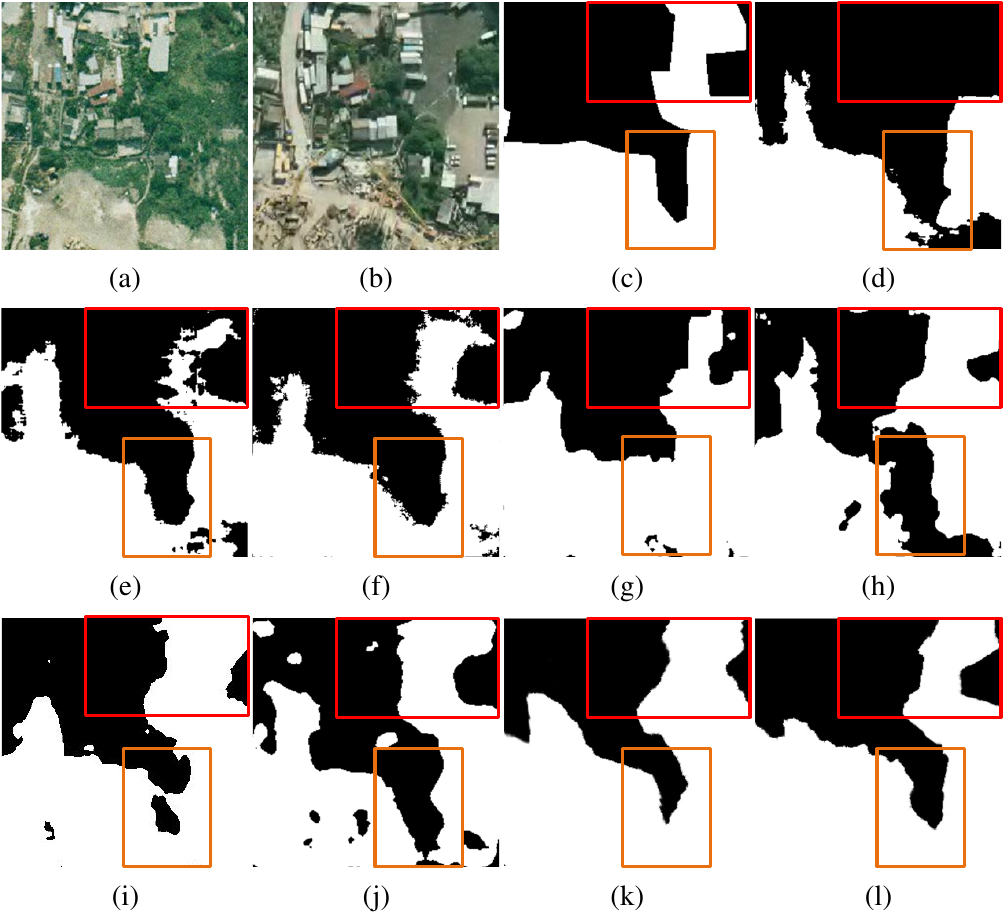} 
    \end{center}
    \caption{Quantitative comparison results of different methods on the SYSU-CD dataset. (a) Image $x_A$. (b) Image $x_B$. (c) Ground truth. (d) FC-EF. (e) FC-Siam-conc. (f) FC-Siam-diff-res. (g) STANet. (h) DSAMNet. (i) ESCNet. (j) DonaNet-base. (k) DonaNet-glw. (l) DonaNet.} 
    \label{fig12}    
\end{figure}

\begin{table}[t]  
  \centering
  \caption{Comparison results of all methods on the SYSU-CD dataset.}
  \resizebox{\linewidth}{!}{
    \begin{tabular}{c|ccccc}
    \toprule
    Method & Pre.(\%) & Rec.(\%) & F1.(\%) & IoU(\%)& OA(\%)  \\
    \midrule
    FC-EF \cite{daudt2018fully} &74.21 &79.41 &76.72 &62.23&89.89  \\
    FC-Siam-conc \cite{daudt2018fully} &76.47 &76.24 &76.35 &61.75&89.85  \\
    FC-Siam-diff-res \cite{daudt2018fully} &76.92 &79.01 &77.96 &63.87&90.31  \\
    FCN-PP \cite{lei2019landslide} &69.81 &76.90 &73.18 &57.71&87.49  \\
    W-Net \cite{hou2019w} &71.08 &78.42 &74.57 &59.45&90.93  \\
    CDGAN \cite{hou2019w} &70.51 &79.03 &74.53 &59.40&89.67  \\
    STANet \cite{chen2020spatial} &70.98 &\underline{81.21} &75.75 &60.97&90.99  \\
    DSAMNet \cite{shi2021deeply} &73.93 &78.31 &76.06 &61.36&91.40  \\
    ESCNet \cite{zhang2021escnet} &80.06 &79.15 &79.60 &66.12 &92.65  \\
    SEIFNet \cite{huang2024spatiotemporal} &78.11 &78.56 &78.33 &64.38 &90.03 \\
    MFINet \cite{ren2024mfinet} &80.36 &79.01 &79.68 &66.22 &91.17 \\
    SAAN \cite{guo2024saan} &81.55 &78.37 &79.93 &66.57 &92.15 \\
    DAMFANet \cite{ren2024dual} &\underline{82.62} &80.31 &\underline{81.45} &\underline{68.70} &\underline{93.70} \\
    \midrule
    DonaNet (Ours) &\textbf{84.79} &\textbf{81.62} &\textbf{83.17} &\textbf{71.20}&\textbf{96.18}  \\
    \bottomrule
    \end{tabular}}
  \label{table3} 
\end{table}

$\bullet$\ \textbf{Quantitative Comparison.} Table \ref{table2} reports the quantitative comparison results. FCN-PP achieves the highest recall while its precision is the lowest, which leads to the reduction in the results on the F1-score and IoU. DSAMNet obtains the second-highest precision and OA, but its F1-score and IoU are non-ideal due to extremely low recall. Moreover, the second-highest F1-score and IoU are achieved by ESCNet. Compared to them, DonaNet achieves the best results on the Precision, F1-score, IoU, and OA, outperforms the second-highest results by $2.75\%$, $2.02\%$, $1.97\%$, and $2.57\%$.

\begin{figure}[t] 
    \setlength{\abovecaptionskip}{-0.4cm}
    \begin{center}
    \centering 
    \includegraphics[width=0.36\textwidth, height=0.32\textwidth]{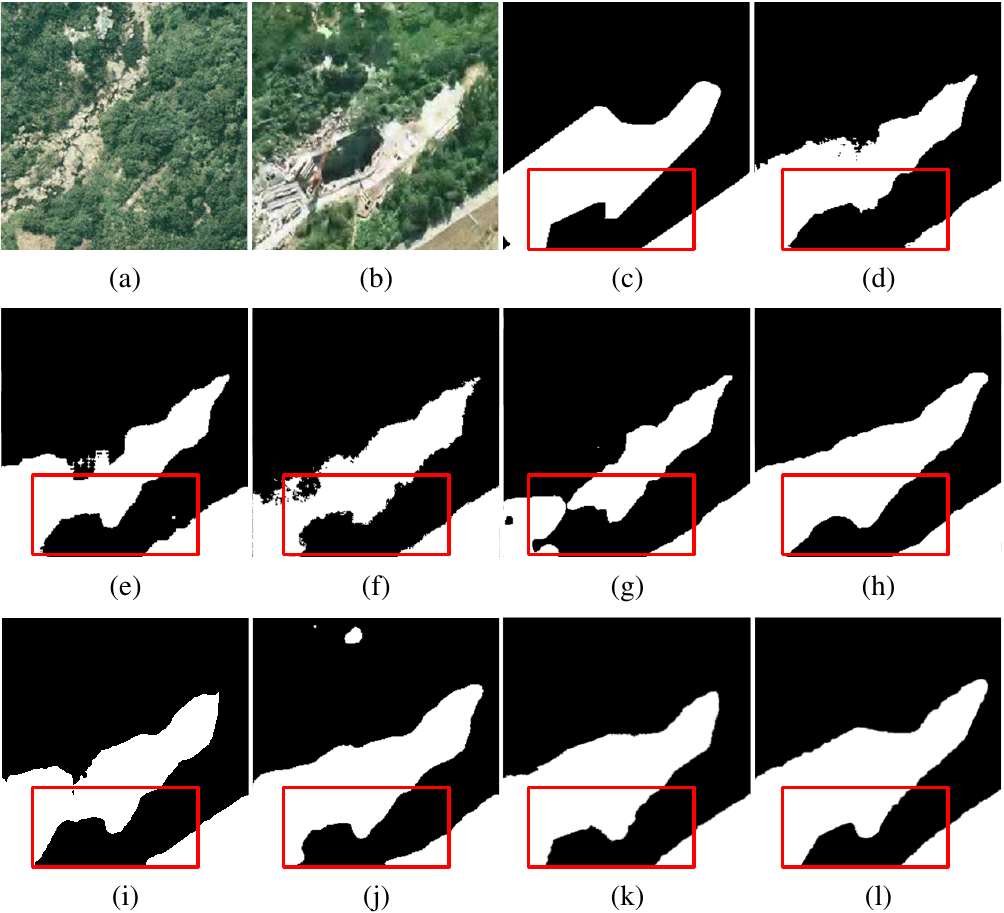} 
    \end{center}
    \caption{Quantitative comparison results of different methods on the SYSU-CD dataset. (a) Image $x_A$. (b) Image $x_B$. (c) Ground truth. (d) FC-EF. (e) FC-Siam-conc. (f) FC-Siam-diff-res. (g) STANet. (h) DSAMNet. (i) ESCNet. (j) DonaNet-base. (k) DonaNet-glw. (l) DonaNet.} 
    \label{fig13}
\end{figure}

\begin{table}[t]  
  \centering
  \caption{Comparison results of all methods on the PCD dataset. $\dagger$ indicates that labels are used during training for fair comparison.}
    \begin{tabular}{c|c|c|c}
    \toprule
    \multirow{2}[4]{*}{Method} & \multirow{2}[4]{*}{Publication} & \multicolumn{2}{c}{F1-score(\%)} \\
    \cmidrule{3-4} &       & Tsunami & GSV \\
    \midrule
    CNN-Feat \cite{jst2015change} & BMVC 2015 & 72.4  & 63.9 \\
    DOF-CDNet \cite{sakurada2017dense}& Arxiv 2017 & 83.8  & 70.3 \\
    CDNet \cite{alcantarilla2018street}& AR 2018 & 77.4  & 61.4 \\
    CosimNet \cite{guo2018learning}& Arxiv 2018 & 80.6  & 69.2 \\
    DASNet \cite{chen2020dasnet}& RS 2020 & 84.1  & 74.5 \\
    CSCDNet \cite{sakurada2020weakly}& ICRA 2020 & 85.9  & 73.8 \\
    HPCFNet \cite{lei2020hierarchical}& TIP 2020 & \underline{86.8}  & 77.6 \\
    SimUNet \cite{zhu2021building}& ARC 2021 & 82.9  & 68.1 \\
    SimSac \cite{park2022dual}& CVPR 2022 & 86.5  & \underline{78.2} \\
    DFMA \cite{lee2024semi}& WACV 2024 &85.1 &76.7 \\
    CCSCD \cite{lee2024color}& MTA 2024 &85.2 &77.0 \\
    ZeroSCD\textsuperscript{$\dagger$} \cite{sundar2024zeroscd}& arXiv 2024 &87.1 &78.9 \\
    ZSSCD\textsuperscript{$\dagger$} \cite{cho2024zero}& arXiv 2024 &87.3 &79.2 \\
    \midrule
    DonaNet (Ours) &\textbf{--} &\textbf{88.7} &\textbf{81.3} \\
    \bottomrule
    \end{tabular}
  \label{ncd}
\end{table}

\begin{table}[t] 
  \centering
  \caption{Comparison results of all methods on the VL-CMU-CD dataset. $\dagger$ indicates that labels are used during training for fair comparison.}
    \begin{tabular}{c|c|c}
    \toprule
    Method & Publication & F1-score(\%) \\
    \midrule
    CNN-Feat \cite{jst2015change} & BMVC 2015 & 40.3 \\
    DOF-CDNet \cite{sakurada2017dense}& Arxiv 2017 &68.8 \\
    CDNet \cite{alcantarilla2018street}& AR 2018 &58.2   \\
    CosimNet \cite{guo2018learning}& Arxiv 2018 &70.6  \\
    DASNet \cite{chen2020dasnet}& RS 2020 & 72.1  \\
    CSCDNet \cite{sakurada2020weakly}& ICRA 2020 & 71.0 \\
    HPCFNet \cite{lei2020hierarchical}& TIP 2020 &75.2   \\
    SimUNet \cite{zhu2021building}& ARC 2021 & 71.4 \\
    SimSac \cite{park2022dual}& CVPR 2022 &\underline{75.6}  \\
    DFMA \cite{lee2024semi}& WACV 2024 &74.3 \\
    CCSCD \cite{lee2024color}& MTA 2024 &74.8 \\
    ZeroSCD\textsuperscript{$\dagger$} \cite{sundar2024zeroscd}& arXiv 2024 &75.3 \\
    ZSSCD\textsuperscript{$\dagger$} \cite{cho2024zero}& arXiv 2024 &76.0 \\
    \midrule
    DonaNet (Ours) &\textbf{--} &\textbf{78.6} \\
    \bottomrule
    \end{tabular}
  \label{ncd1}
\end{table}

\begin{figure*}[t] 
    \setlength{\abovecaptionskip}{-0.35cm}
    \begin{center}
    \centering 
    \includegraphics[width=0.8\textwidth, height=0.45\textwidth]{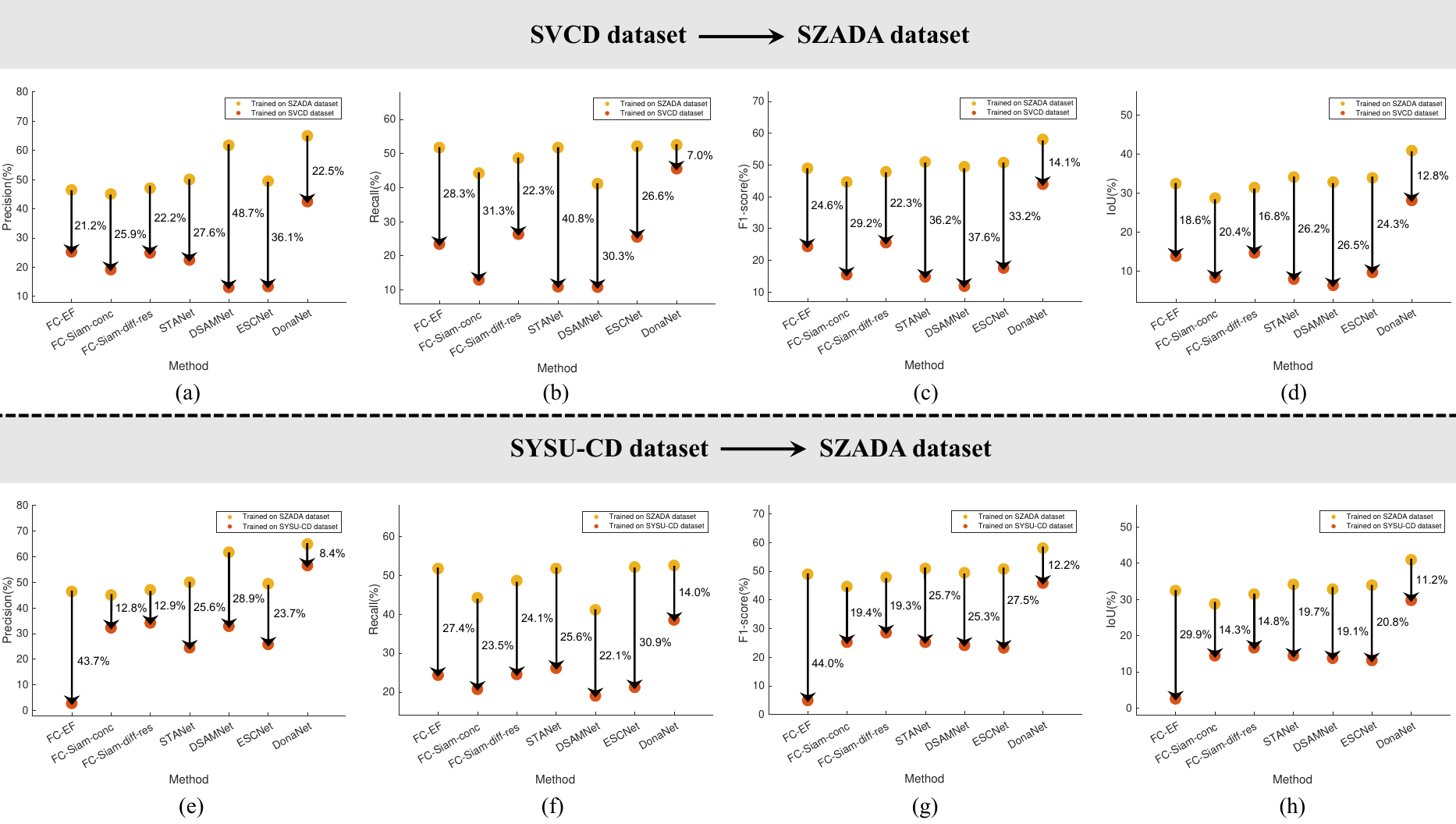} 
    \end{center}
    \caption{Comparison results on two cross-scene settings. The orange dots in the first/second row are the results obtained using a model trained on the SVCD/SYSU-CD to predict the unseen SZADA. The yellow dots in these two rows are the results obtained using the model trained on the SZADA to predict the SZADA. The fluctuation range is retained to one decimal place.} 
    \label{fig11}   
\end{figure*}

\begin{figure}[t] 
    \setlength{\abovecaptionskip}{-0.20cm}
    \begin{center}
    \centering 
    \includegraphics[width=0.49\textwidth, height=0.25\textwidth]{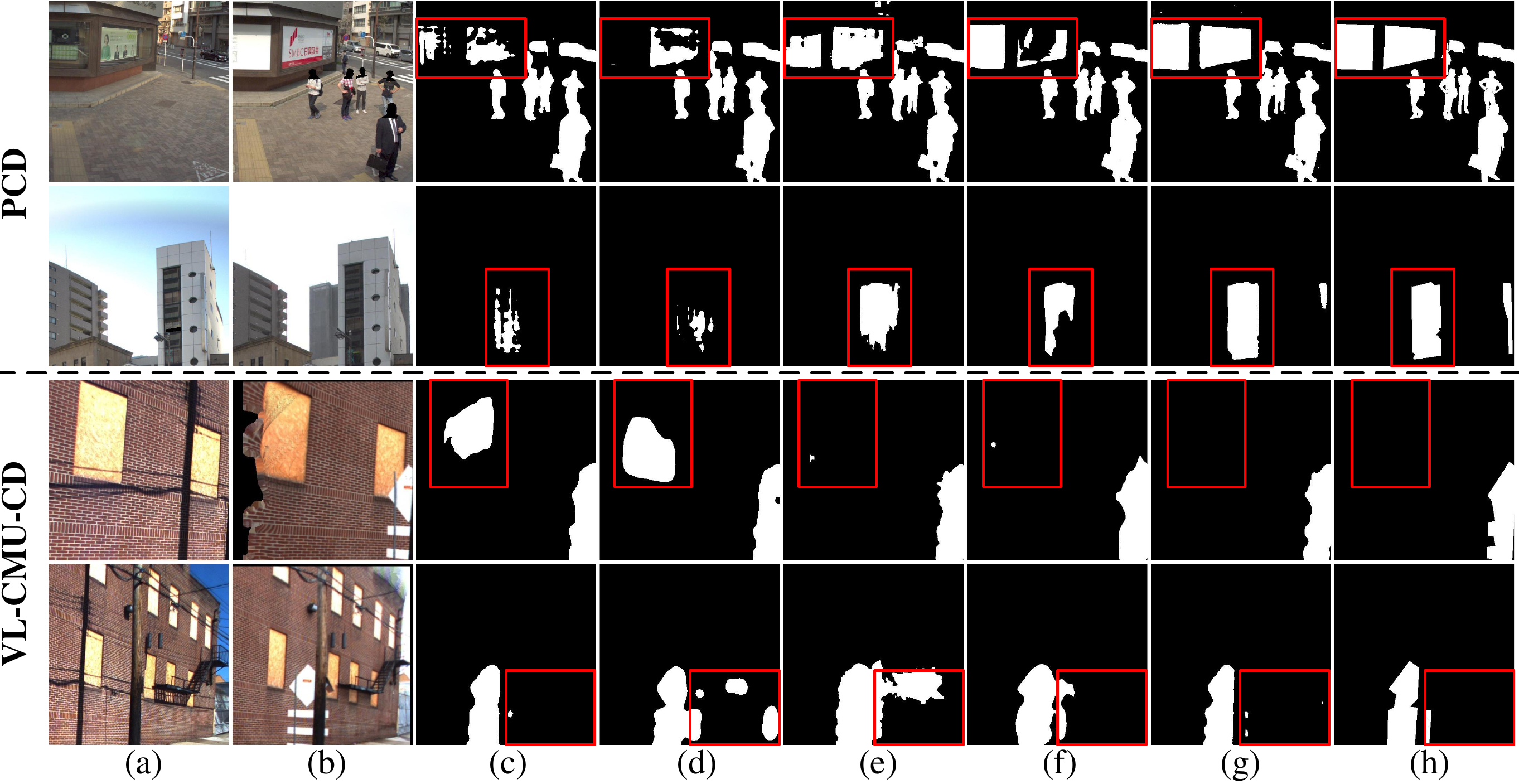} 
    \end{center}
    \caption{Quantitative comparison results of different methods on the PCD and VL-CMU-CD datasets. (a) Image $x_A$. (b) Image $x_B$. (c) HPCFNet. (d) SimUNet. (e) SimSac. (f) DFMA. (g) DonaNet. (h) Ground truth.}
    \label{na}
\end{figure}

\subsubsection{Comparison on the SYSU-CD Dataset}
$\bullet$\ \textbf{Qualitative Comparison.} As discussed in Section \ref{4.1}, unlike the SVCD and SZADA datasets, this dataset contains more diverse changed instances, such as urban buildings, sea construction, road expansion, etc. Thus, the above facts bring more significant challenges. To qualitatively demonstrate the performance of different methods for these diverse change types, we show the visual comparison results of each method on the SYSU-CD dataset in Fig. \ref{fig12} and Fig. \ref{fig13}. From Fig. \ref{fig12}, it can be seen that the types of changes include newly built urban buildings and road expansion simultaneously. In Fig. \ref{fig12}(d)-(f), the detection results obtained by the FC-EF, FC-Siam-conc, and FC-Siam-diff-res methods have rough and jagged edges, and they can only detect the main changing regions. The results of the other three existing methods (STANet, DSAMNet, and ESCNet) shown in Fig. \ref{fig12}(g)-(i), although the edges are relatively smooth, are over-detected for many pixels and the contours are not precise enough. For the sample in Fig. \ref{fig13}, the type of change is groundwork before construction. As shown in Fig. \ref{fig13}(d)-(i), except for the unsmooth edges detected by these existing methods, there are apparent omissions for pixels that are actually changed. In comparison, our proposed method DonaNet has the best visual effect on these types of changes and detects the contours of changed objects more accurately, as shown in Fig. \ref{fig12}(l) and Fig. \ref{fig13}(l). Besides, qualitative comparisons for pseudo-changes are analyzed in detail,
\noindent\textbf{Analysis of Pseudo-changes:} Due to style differences caused by variations in illumination/weather, existing methods fail to escape pseudo-changes when detecting. For the area in the orange rectangular box in Fig. \ref{fig12}, the herbaceous plants from different phases have slight differences in color. They suffer from overdetection in this region, especially STANet and ESCNet, as shown in Fig. \ref{fig12}(g) and Fig. \ref{fig12}(i). 
It's notable that several methods partially missed detection in this area. Upon analysis, we find that they are misled by similar appearances, failing to adequately capture the essential semantic content. In contrast, after adding the domain difference removal module and the cross-temporal generalization learning, DonaNet reduces the overdetection.

$\bullet$\ \textbf{Quantitative Comparison.} Table \ref{table3} further reports the quantitative results. STANet achieves the second highest recall, but its precision is very low, indicating that many unchanged pixels are wrongly detected. This is consistent with the qualitative results analyzed above. ESCNet achieves relatively better performance, with the second-highest precision, F1-score, IoU, OA, and KC. Among these methods, DonaNet achieves the best results on Precision, Recall, F1-score, IoU, and OA metrics, which surpass the second-highest results by $2.17\%$, $0.41\%$, $1.72\%$, $2.50\%$, and $2.48\%$. The superior performance obtained under complex scenes further demonstrates the better generalization ability of our method while embodying that it is crucial to eliminate domain shift within bitemporal image pairs.

\subsubsection{Comparison on the Natural Datasets}
To further validate the generalizability of our method, we conducted experiments on two widely used natural change detection datasets, PCD \cite{jst2015change} and VL-CMU-CD \cite{alcantarilla2018street}, following standard evaluation protocols in the field \cite{lei2020hierarchical,guo2023local,sakurada2020weakly,sakurada2017dense} and using F1-score as the performance metric. The quantitative results in Tables \ref{ncd}-\ref{ncd1} demonstrate that our method outperforms state-of-the-art techniques on these datasets. Additionally, the qualitative results in Fig. \ref{na} highlight the superior performance of our method in detecting changes, particularly in scenarios with significant style shifts (e.g., changes in lighting or weather), showcasing its robustness for natural image change detection.

\subsection{Generalization on Cross-scene Image Pairs} \label{4}
To demonstrate the generalizability of our method DonaNet, we experiment with DonaNet and six other existing methods on cross-scene image pairs. Specifically, we use the model trained on the SVCD or SYSU-CD datasets to make predictions directly on the unseen test dataset SZADA. To illustrate more intuitively, we compare the above two cross-scene settings with the results obtained by the model trained on the SZADA dataset, and the performance changes are reported in Fig. \ref{fig11}(a)-(d) and Fig. \ref{fig11}(e)-(h) via multiple scatterplots. 

\begin{table}[t] 
  \centering
  \caption{Comparative results of testing models across multiple scenes. M-SVCD/M-SYSU/M-SZADA indicates the model trained on the SVCD/SYSU/SZADA dataset. The fluctuation range is retained to one decimal place.}
  \resizebox{\linewidth}{!}{
    \begin{tabular}{c|cc|cc|cc|cc}
    \toprule
    \multirow{2}[4]{*}{Method} &\multicolumn{4}{c|}{F1-score(\%)} & \multicolumn{4}{c}{IoU(\%)} \\
    \cmidrule{2-9} & M-SVCD & M-SYSU & M-SZADA & M-SYSU & M-SVCD & M-SYSU & M-SZADA & M-SYSU \\
    \midrule
    FC-EF \cite{daudt2018fully} &82.46 &51.83 \textcolor[RGB]{255,0,0}{($\downarrow$30.6)}&48.92 &4.90 \textcolor[RGB]{255,0,0}{($\downarrow$44.0)} &71.15 &34.98 \textcolor[RGB]{255,0,0}{($\downarrow$36.2)}&32.38 &2.51 \textcolor[RGB]{255,0,0}{($\downarrow$29.9)} \\
    FC-Siam-conc \cite{daudt2018fully} &80.22 &47.05 \textcolor[RGB]{255,0,0}{($\downarrow$33.2)}&44.57 &25.16 \textcolor[RGB]{255,0,0}{($\downarrow$19.4)} &67.98 &30.76 \textcolor[RGB]{255,0,0}{($\downarrow$37.2)}&28.67 &14.39 \textcolor[RGB]{255,0,0}{($\downarrow$14.3)} \\
    FC-Siam-diff-res \cite{daudt2018fully} &87.80 &57.16 \textcolor[RGB]{255,0,0}{($\downarrow$30.6)}&47.82 &28.55 \textcolor[RGB]{255,0,0}{($\downarrow$19.3)} &79.25 &40.02 \textcolor[RGB]{255,0,0}{($\downarrow$39.2)}&31.42 &16.65 \textcolor[RGB]{255,0,0}{($\downarrow$14.8)} \\
    STANet \cite{chen2020spatial} &91.33 &60.20 \textcolor[RGB]{255,0,0}{($\downarrow$31.1)}&50.86 &25.21 \textcolor[RGB]{255,0,0}{($\downarrow$25.7)} &84.04 &43.06 \textcolor[RGB]{255,0,0}{($\downarrow$41.0)}&34.10 &14.42 \textcolor[RGB]{255,0,0}{($\downarrow$19.7)} \\
    DSAMNet \cite{shi2021deeply} &92.90 &66.22 \textcolor[RGB]{255,0,0}{($\downarrow$26.7)}&49.36 &24.09 \textcolor[RGB]{255,0,0}{($\downarrow$25.3)} &86.60 &49.50 \textcolor[RGB]{255,0,0}{($\downarrow$37.1)}&32.77 &13.70 \textcolor[RGB]{255,0,0}{($\downarrow$19.1)} \\
    ESCNet \cite{zhang2021escnet} &92.83 &55.80 \textcolor[RGB]{255,0,0}{($\downarrow$37.0)}&50.68 &23.22 \textcolor[RGB]{255,0,0}{($\downarrow$27.5)} &86.63 &38.70 \textcolor[RGB]{255,0,0}{($\downarrow$47.9)}&33.94 &13.14 \textcolor[RGB]{255,0,0}{($\downarrow$20.8)} \\
    \midrule
    DonaNet (Ours) &96.36 &85.16 \textcolor[RGB]{255,0,0}{($\downarrow$11.2)}&58.04 &45.80 \textcolor[RGB]{255,0,0}{($\downarrow$12.2)} &92.98 &74.16 \textcolor[RGB]{255,0,0}{($\downarrow$18.8)}&40.88 &29.70 \textcolor[RGB]{255,0,0}{($\downarrow$11.2)} \\
    \bottomrule
    \end{tabular}}
  \label{tab:addlabel}
\end{table}

As shown in Fig. \ref{fig11}, due to the domain shift between different scenes, the performance of existing methods trained on SVCD and SYSU-CD is significantly degraded on SZADA. From Fig. \ref{fig11}(a) and Fig. \ref{fig11}(e), we can see that most of the methods achieve higher precision by using the model trained on SYSU-CD than the model trained on SVCD, \emph{e.g.}, FC-Siam-conc, FC-Siam-diff-res, STANet, DSAMNet, and ESCNet. We infer that this occurs because the SYSU-CD dataset covers more diverse instances of change than the SVCD dataset. Hence, the domain differences between SYSU-CD and SZADA is smaller than that between SVCD and SZADA, restraining the extent to which the model's generalization is harmed. Furthermore, in Fig. \ref{fig11}(b) and Fig. \ref{fig11}(f), existing methods achieve slightly higher recall on the second cross-scene setting (SYSU-CD$\rightarrow$SZADA) than on the first cross-scene setting (SVCD$\rightarrow$SZADA), which limits their F1-score and IoU results on the second setting. In contrast, our method DonaNet is relatively less degenerate on both settings due to the learned domain-agnostic difference representation. Compared to existing methods, DonaNet finally achieves the best results on the four evaluation metrics in both settings.

To further assess its robustness across various domain shifts, we extend the experimental setup by training the model on one dataset (e.g., SYSU-CD) and testing it on multiple others (e.g., SVCD and SZADA). The results are presented in Table \ref{tab:addlabel}. The results show that, while all change detection methods experience performance degradation due to domain shift, our method demonstrates less degradation compared to others, validating its robustness and superior ability to mitigate performance loss in cross-domain scenarios. These findings provide strong evidence of the generalization capabilities of our method in real-world, heterogeneous datasets.

\begin{table}[t] 
  \centering
  \caption{Our method is incorporated into existing state-of-the-art methods for change detection, respectively. Quantitative results are reported on the SVCD and SYSU-CD datasets.}
    \resizebox{\linewidth}{!}{
    \begin{tabular}{c|cccc|cccc}
    \toprule
    \multirow{2}[4]{*}{Method} & \multicolumn{4}{c|}{SVCD}& \multicolumn{4}{c}{SYSU-CD} \\
    \cmidrule{2-9} & Pre.(\%) & Rec.(\%) & F1(\%) & IoU(\%)& Pre.(\%) & Rec.(\%) & F1(\%) & IoU(\%) \\
    \midrule
    \tabincell{c}{FC-EF \cite{daudt2018fully}\\ +Ours} &\tabincell{c}{77.48\\ \textbf{93.73}}& \tabincell{c}{88.13\\ \textbf{92.69}}& \tabincell{c}{82.46\\ \textbf{93.21}}& \tabincell{c}{71.15\\ \textbf{87.28}}& \tabincell{c}{74.21\\ \textbf{83.69}}& \tabincell{c}{79.41\\ \textbf{81.40}}& \tabincell{c}{76.72\\ \textbf{82.53}}& \tabincell{c}{62.23\\ \textbf{70.25}} \\
    \midrule
    \tabincell{c}{FC-Siam-conc \cite{daudt2018fully}\\ +Ours} &\tabincell{c}{72.88\\ \textbf{95.11}}& \tabincell{c}{89.21\\ \textbf{93.48}}& \tabincell{c}{80.22\\ \textbf{94.29}}& \tabincell{c}{67.98\\ \textbf{89.19}}& \tabincell{c}{76.47\\ \textbf{84.67}}& \tabincell{c}{76.24\\ \textbf{80.63}}& \tabincell{c}{76.35\\ \textbf{82.60}}& \tabincell{c}{61.75\\ \textbf{70.36}}  \\
    \midrule
    \tabincell{c}{FCN-PP \cite{lei2019landslide}\\ +Ours} &\tabincell{c}{83.05\\ \textbf{95.39}}&\tabincell{c}{91.80\\ \textbf{95.86}}&\tabincell{c}{87.21\\ \textbf{95.62}}&\tabincell{c}{77.31\\ \textbf{91.62}}& \tabincell{c}{69.81\\ \textbf{84.11}}&\tabincell{c}{76.90\\ \textbf{81.09}}&\tabincell{c}{73.18\\ \textbf{82.57}}&\tabincell{c}{57.71\\ \textbf{70.32}}  \\
    \midrule
    \tabincell{c}{STANet \cite{chen2020spatial}\\ +Ours} &\tabincell{c}{89.24\\ \textbf{96.98}}&\tabincell{c}{93.51\\ \textbf{95.95}}&\tabincell{c}{91.33\\ \textbf{96.46}}&\tabincell{c}{84.04\\ \textbf{93.17}}& \tabincell{c}{70.98\\ \textbf{85.31}}&\tabincell{c}{81.21\\ \textbf{82.06}}&\tabincell{c}{75.75\\ \textbf{83.65}}&\tabincell{c}{60.97\\ \textbf{71.90}} \\
    \midrule
    \tabincell{c}{DSAMNet \cite{shi2021deeply}\\ +Ours} &\tabincell{c}{93.35\\ \textbf{97.19}}&\tabincell{c}{92.41\\ \textbf{96.52}}&\tabincell{c}{92.90\\ \textbf{96.85}}&\tabincell{c}{86.60\\ \textbf{93.90}}& \tabincell{c}{73.93\\ \textbf{85.20}}&\tabincell{c}{78.31\\ \textbf{81.11}}&\tabincell{c}{76.06\\ \textbf{83.10}}&\tabincell{c}{61.36\\ \textbf{71.09}} \\
    \bottomrule
    \end{tabular}}
  \label{table4} 
\end{table}

\subsection{Applicability of Our Method to Existing Methods}
Obviously, our method DonaNet is complementary to existing state-of-the-art methods for change detection, and it can be easily incorporated into these methods to boost their performance further. To validate the above conclusion, we incorporate the domain difference removal (DDR) and cross-temporal generalization learning (CTGL) modules in DonaNet into five state-of-the-art change detection methods. Table \ref{table4} reports the quantitative results on two datasets SVCD and SYSU-CD. As shown in Table \ref{table4}, incorporating our method further consistently improves the performance of all state-of-the-art methods. The positive effect of DonaNet can be attributed to assisting other models in avoiding overfitting domain-specific style information. By adding our strategies, the generalization of features can be further improved to boost detection performance. Moreover, note that after incorporating our method, the inference process is not introduced with a lot of extra parameters and computation once the model is trained.

\begin{table}[t] 
  \centering
  \caption{Ablation studies of our proposed method on the SVCD dataset.}
    \resizebox{\linewidth}{!}{
    \begin{tabular}{c|ccccc|cccc}
    \toprule
    Model &\tabincell{c}{DDR\\-gln} &\tabincell{c}{DDR\\-glw} &\tabincell{c}{CTGL\\-UST} &\tabincell{c}{CTGL\\-BST} &\tabincell{c}{CTGL\\-IBST} & Pre.(\%) & Rec.(\%) & F1(\%) & IoU(\%) \\
    \midrule
    DonaNet-unweighted & &  &  &  &  &93.24 &93.86 &93.55 &87.88 \\
    DonaNet-base & &  &  &  &  &93.71 &94.35 &94.03 &88.73 \\
    \midrule
    DonaNet-gln &$\checkmark$ &  &  &  &  &96.02 &95.23 &95.62 &91.61 \\
    DonaNet-glw & &$\checkmark$ &  &    &  &96.37 &95.29 &95.83 &91.99 \\
    DonaNet-mst\_v1 & & &$\checkmark$ & & &95.38 &94.65 &95.01 &90.50  \\
    DonaNet-mst\_v2 & & & &$\checkmark$ & &95.42 &94.63 &95.02 &90.52   \\
    DonaNet-mst\_v3 & & & & &$\checkmark$ &95.48 &94.61 &95.04 &90.55   \\
    DonaNet-mst\_v4 & & &$\checkmark$ &$\checkmark$ & &95.70 &94.75 &95.22 &90.88   \\
    DonaNet-mst\_v5 & & &$\checkmark$ & &$\checkmark$ &95.80 &94.82 &95.31 &91.04   \\
    DonaNet-mst\_v6 & & & &$\checkmark$ &$\checkmark$ &95.78 &94.81 &95.29 &91.01   \\
    DonaNet-mst &  &   &$\checkmark$ &$\checkmark$ &$\checkmark$ &95.89 &94.93 &95.41 &91.22 \\
    DonaNet-gln-mst &$\checkmark$ &  &$\checkmark$ &$\checkmark$ &$\checkmark$ &\underline{96.71} &\underline{95.69} &\underline{96.20} &\underline{92.67} \\
    DonaNet &  &$\checkmark$ &$\checkmark$ &$\checkmark$ &$\checkmark$ &\textbf{96.93} &\textbf{95.80} &\textbf{96.36} &\textbf{92.98} \\
    \bottomrule
    \end{tabular}}
  \label{table5} 
\end{table}

\subsection{Ablation Studies}
\subsubsection{Performance Impact of Each Module}
To evaluate each component, we conduct experiments with different combinations of these components on the SVCD dataset. Table \ref{table5} shows the experimental conditions and reports results.

As shown in Table \ref{table5}, the SD network (DonaNet-base) without arbitrary modules achieves $93.71\%$, $94.35\%$, $94.03\%$, and $88.73\%$ on the precision, recall, F1-score, and IoU, slightly higher than the network without assigning specific weights to the samples (DonaNet-unweighted). Adding the gln layer in the DDR module to the DonaNet-base improves the performance on four metrics by $2.31\%$, $0.88\%$, $1.59\%$, and $2.88\%$. Replacing the gln layer with the glw layer brings even more performance gains. By combining the three transformation modes (UST + BST + IBST) in the CTGL module, a multi-style transformation (mst) is formed. Adding the CTGL module with mst (CTGL-mst) improves performance to a lesser extent than adding the DDR module, which is improved by $2.18\%$, $0.58\%$, $1.38\%$, and $2.49\%$ on these metrics. We also report the performance impact of adding different combinations of transformations (mst\_v1-mst\_v6) to the DonaNet-base. The detection performance is better when more transformations are used because the model can learn more enhanced data (broaden the training distribution) by various transformations, which is conducive to the learning of generalization ability. DonaNet-gln by adding the CTGL-mst module further boosts performance of the network. Finally, adding the CTGL-mst module to DonaNet-glw achieves the best performance on evaluation metrics, \emph{i.e.}, $96.93\%$, $95.80\%$, $96.36\%$, and $92.98\%$. Thus, we can conclude that the DDR and CTGL are complementary and each module contributes to these improvements. We report the combination with the best performance among all other experiments, calling it DonaNet.

\begin{figure}[t] 
    \setlength{\abovecaptionskip}{-0.35cm}
    \begin{center}
    \centering 
    \includegraphics[width=0.41\textwidth, height=0.22\textwidth]{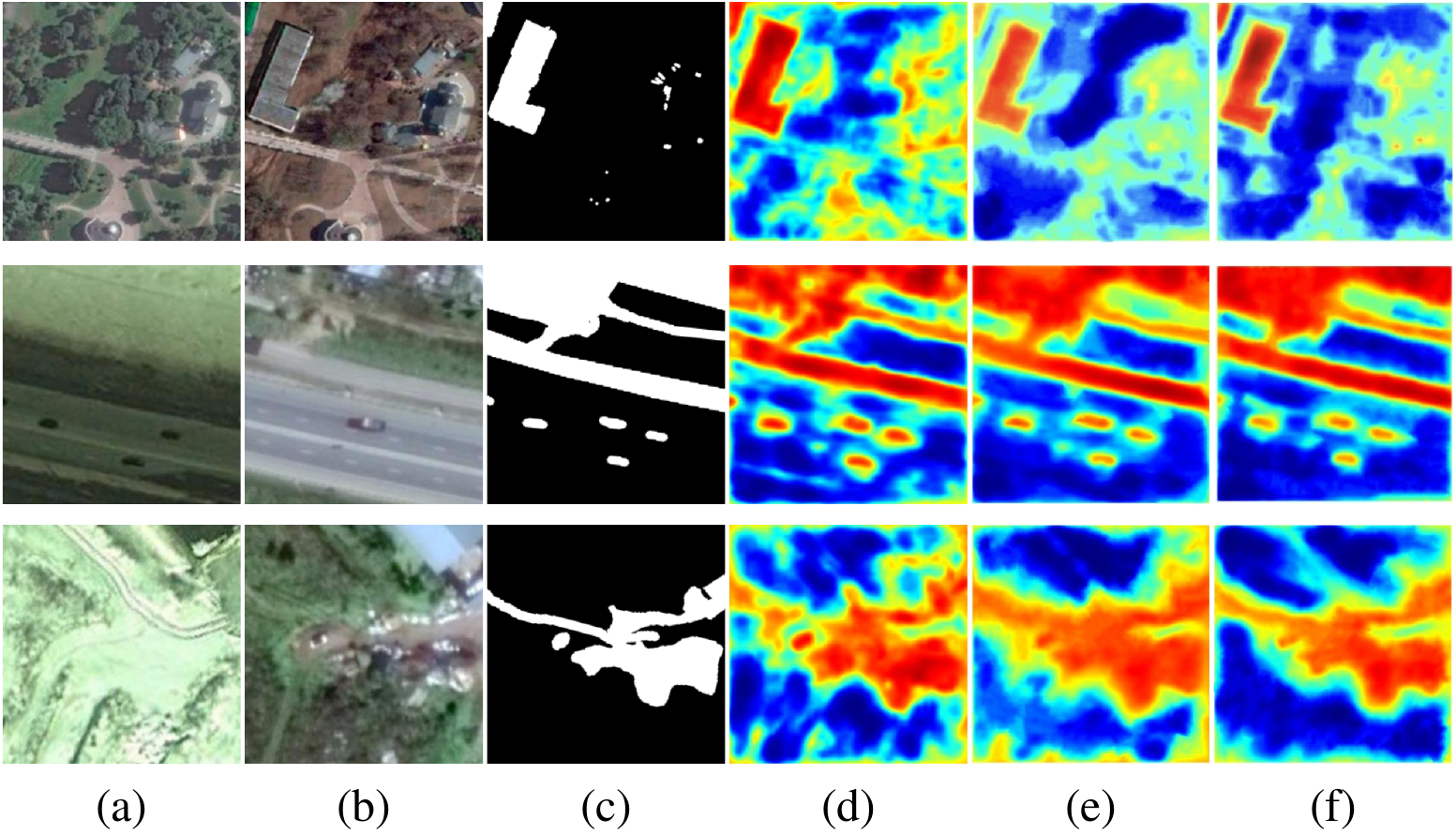} 
    \end{center}
    \caption{Visualization of the effects of DDR and CTGL modules. (a) Image $x_A$. (b) Image $x_B$. (c) Ground truth. (d) Features without the DDR and CTGL modules. (e) Features without the CTGL module. (f) Features with the DDR and CTGL modules.} 
    \label{fig9}    
\end{figure}

\subsubsection{Effects of DDR and CTGL Modules}
To verify the effectiveness of the DDR and CTGL modules, we visualize the results for three bitemporal image pairs on the SVCD dataset, including without the DDR and CTGL modules, with the DDR module, and with the DDR and CTGL modules. The features are averaged and normalized in the channel dimension to obtain visualization results, as shown in Fig. \ref{fig9}. The red regions indicate higher attention, and the blue regions indicate lower attention. From Fig. \ref{fig9}(d)-(e), the network with the DDR module can more accurately detect the contour of the changed objects while reducing the focus on the non-changed areas. By further adding the CTGL module, the edges of the detection results are more refined and fit the actual changes in the images. In the second row of Fig. \ref{fig9}(f), narrow and slightly curved roads are ideally detected, and the network ignores the noisy regions due to style differences in the first and third rows to improve the detection accuracy.

\begin{table}[t] 
  \centering
  \caption{Comparison of computational efficiency of different methods on the SVCD dataset.}
    \resizebox{\linewidth}{!}{
    \begin{tabular}{c|cccccccccc}
    \toprule
    Method &FC-EF &\tabincell{c}{FC-Siam\\-conc} &\tabincell{c}{FC-Siam-\\diff-res} &FCN-PP &W-Net &CDGAN &STANet &DSAMNet &ESCNet & DonaNet \\
    \midrule
    Params.(MB) &1.35 &1.55 &1.35 &27.81 &40.49 &115.12 &16.93 &17.00 &5.12 &3.80 \\
    FLOPs(GB) &2.68 &4.06 &3.50 &34.81 &94.89 &164.74 &14.40 &37.02 &11.65 &11.48 \\
    Time(ms) &13.04 &13.58 &13.56 &30.49 &16.28 &51.91 &41.28 &58.03 &130.97 &15.05 \\
    \bottomrule
    \end{tabular}}
  \label{table6}  
\end{table}

\subsection{Comparison and Discussion of Efficiency} 
Our method aims to achieve high-quality detection by a model with few parameters. To analyze and compare the computational efficiency of our method with existing methods, we record the number of parameters (Params.), floating-point operations per second (FLOPs), and inference time (Time) of each method. For a fair comparison, all methods are reproduced and tested on a new server equipped with an Nvidia Tesla RTX3090 GPU with 24G memory. The results are reported in Table \ref{table6}, where Time is the average inference time calculated on 100 randomly selected images of size $256 \times 256 \times 3$. As shown in the first row of Table \ref{table6}, except for the ESCNet method, the recently proposed methods with better performance are all heavily parameterized, \emph{e.g.}, CDGAN has the most trainable parameters of 115.12 MB due to its large number of convolution operations. In the second and third rows of Table \ref{table6}, it can be seen that the computational costs of these recent methods are also correspondingly large. In comparison, our method DonaNet has the smallest trainable parameters other than the previous FC-EF, FC-Siam-conc and FC-Siam-diff-res methods, \emph{i.e.}, only 3.80 MB. It is worth mentioning that the two modules DDR and CTGL designed in our method do not contain trainable parameters, so they do not bring any overhead regarding the model size. Furthermore, our method has less computational cost (14.75 ms) while maintaining the best detection performance.

\subsection{Discussion of Hyper-parameter Sensitivity}
To give guidance for the practical applications of the proposed method, we discuss the hyper-parameter setting for dividing local regions in the style proxy.
As specified in Section \ref{c} and \ref{d}, we employ channel-wise statistics of local-level features/images to proxy style. The number of local regions in a feature or image is controlled by the hyper-parameters $\lambda$ and $\lambda'$, respectively. The number of local regions is proportional to $\lambda$ and $\lambda'$. We study the sensitivity of $\lambda$ and $\lambda'$ on the SVCD dataset by changing them from 2 to 16. The experimental results are reported in Table \ref{table7}. As shown in the Table, an appropriate number of local regions (\emph{i.e.}, appropriately sized each region) can better characterize the style to bring detection accuracy. However, when there are too few or too many local regions, the performance of the model decreases. Both of the above are not conducive to the model learning representations that are robust to domain shifts. Furthermore, the optimal $\lambda$ for the control feature is larger than the optimal $\lambda'$ for the control image. We analyze that this is because the downsampled feature map aggregates object information, so smaller local regions can ensure the integrity and independence of the semantic information of the objects contained therein. This indicates the necessity of setting reasonable optimal hyper-parameters $\lambda$ and $\lambda'$.

\begin{table}[t]
  \centering
  \caption{Comparison on the SVCD dataset according to $\lambda$/$\lambda'$ value.}
  \resizebox{\linewidth}{!}{
    \begin{tabular}{c|ccccc|ccccc}
    \toprule
    \multirow{2}[4]{*}{Model} &\multicolumn{5}{c|}{Feature Space} &\multicolumn{5}{c}{Image Space} \\
    \cmidrule{2-11}&$\lambda$=2&$\lambda$=4&$\lambda$=6&$\lambda$=8&$\lambda$=10&$\lambda'$=2&$\lambda'$=4&$\lambda'$=6&$\lambda'$=8&$\lambda'$=10 \\
    \midrule
    Pre.(\%)&95.51&96.15&\textbf{96.93}&95.98&95.32&95.30&95.65&96.27&\textbf{96.93}&95.28 \\
    Rec.(\%)&94.27&95.09&\textbf{95.80}&95.36&95.19&94.21&94.70&95.21&\textbf{95.80}&95.01 \\
    F1(\%)&94.89&95.62&\textbf{96.36}&95.67&95.25&94.75&95.17       &95.74&\textbf{96.36}&95.14 \\
    IoU(\%)&90.27&91.60&\textbf{92.98}&91.70&90.94&90.03&90.79       &91.82&\textbf{92.98}&90.74 \\
    \bottomrule
    \end{tabular}}
  \label{table7}
\end{table}

\section{Limitations}
While our method leverages channel-wise statistics as proxies for style to learn domain-agnostic representations, we acknowledge that these proxies may not fully capture all style variations, especially in more complex datasets. To address this, we plan to explore more refined proxy criteria, such as deep feature-based style estimation or multi-scale feature extraction. Additionally, the whitening operation used to remove style information introduces computational overhead, limiting efficiency for large-scale or real-time applications. Future work will investigate more computationally efficient alternatives, such as attention-based mechanisms, to reduce this cost. Despite these limitations, we believe that the core idea of masking style information to learn domain-agnostic representations offers a promising direction for future research, enhancing model robustness and inspiring advancements in change detection.

\section{Conclusion}
This paper focuses on the problem of pseudo-changes in change detection (CD). Oriented from pseudo-changes caused by style differences, we argue for local-level statistics as style proxies and present a generalizable domain-agnostic difference learning network, called DonaNet. In DonaNet, two approaches are proposed to learn domain-agnostic representations: (1) We design global-to-local normalization (\emph{gln}) to remove domain-specific style information while preserving the discriminability of encoded features. Also, we replace normalization in \emph{gln} with whitening to enhance the generalization of feature representations. (2) We perform style transformation on samples form arbitrary image pairs in three modes to learn feature representations robust to domain shifts. To eliminate the prediction disagreement of the network for the pre- and post-transformed samples, an explicit constraint is added to achieve alignment. Sufficient experiments are conducted on three publicly available change detection datasets to demonstrate the effectiveness of our method. Compared to existing state-of-the-art methods, DonaNet achieves superior qualitative and quantitative results with a smaller model size.




\ifCLASSOPTIONcaptionsoff
  \newpage
\fi



%

\bibliographystyle{IEEEtran}
\bibliography{IEEEabrv,my}







\end{document}